\documentclass[10pt,twocolumn,letterpaper]{article}

\PassOptionsToPackage{table}{xcolor}
\usepackage[algorithms,pagenumbers]{wacv}

\usepackage{booktabs,tabularx,siunitx,multirow}
\usepackage{algorithm}
\usepackage{amsmath}        
\usepackage{amssymb}        

\usepackage{tikz}
\usepackage{colortbl}       
\usepackage{graphicx}       
\usepackage{algpseudocode}
\makeatletter
\providecommand{\theHALG@line}{}
\renewcommand{\theHALG@line}{\thealgorithm.\arabic{ALG@line}}
\makeatother
\usepackage{arydshln}

\definecolor{lightgreen}{RGB}{220, 245, 220}
\definecolor{midgreen}{RGB}{0, 150, 0}

\usepackage[most]{tcolorbox} 
\usepackage{enumitem}        
\usepackage{cuted}
\usepackage{afterpage}

\definecolor{revSJ2h}{rgb}{0.80,0.10,0.10}   
\definecolor{revSzfy}{rgb}{0.10,0.50,0.10}   
\definecolor{revF7qM}{rgb}{0.10,0.30,0.70}   

\definecolor{wacvblue}{rgb}{0.21,0.49,0.74}
\usepackage[pagebackref,breaklinks,colorlinks,allcolors=wacvblue]{hyperref}

\title{PromptPack: Training-Free Augmentation of Diffusion Editors for Intent-Aligned Spatial and Semantic Control}

\author{\makebox[0pt][c]{{\fontsize{12}{14}\selectfont Sophia Zalewski\hspace{0.7em}Mustafa Munir\hspace{0.7em}Enting Liu\hspace{0.7em}Anjul Patney\hspace{0.7em}David Dynerman}}\\[10pt]
NVIDIA\\[4pt]
{\ttfamily\small\href{mailto:szalewski@nvidia.com}{szalewski@nvidia.com}}}

\hypersetup{
  pdftitle={PromptPack: Training-Free Augmentation of Diffusion Editors for Intent-Aligned Spatial and Semantic Control},
  pdfauthor={Sophia Zalewski, Mustafa Munir, Enting Liu, Anjul Patney, David Dynerman}
}

\begin{document}
\maketitle
\begin{abstract}
Many image-editing tasks require confining changes to intended spatial regions while exactly preserving user-designated source content. Instruction-driven editing pipelines often leave such user-intent underspecified, while inpainting editors condition generation on explicitly defined edit support but not on the spatial assignment of individual edit concepts. PromptPack addresses both limitations by equipping diffusion editors with two explicit `Edit Guides': an edit-role partition that labels each pixel as `Keep,' `Permit,' or `Modify' to specify \textbf{\textup{where}} edits may occur, and a regional-prompt assignment that maps spatial regions to text prompts to specify \textbf{\textup{which}} concepts apply where. These guides may be manually specified or, for source-object edits, automatically derived from a text-grounded SAM~3 mask. Without retraining, PromptPack injects independently embedded regional prompts into the editor's fixed-length text sequence and uses the Edit Guides to spatially mask attention, gate flow updates, and restore protected source pixels. We integrate PromptPack into FlowEdit FLUX, FlowEdit SD3, and FLUX Kontext and demonstrate improved requested-concept placement and protected-content preservation across all three integrations on PIE-Bench++. In a pairwise user study, PromptPack achieves an \textbf{87\% preference rate} over FlowEdit FLUX, establishing PromptPack as a practical, training-free interface for intent-aligned diffusion editing.
\end{abstract}

\section{Introduction}
\label{sec:intro}

\begin{figure}[t]
  \centering
  \includegraphics[width=\linewidth]{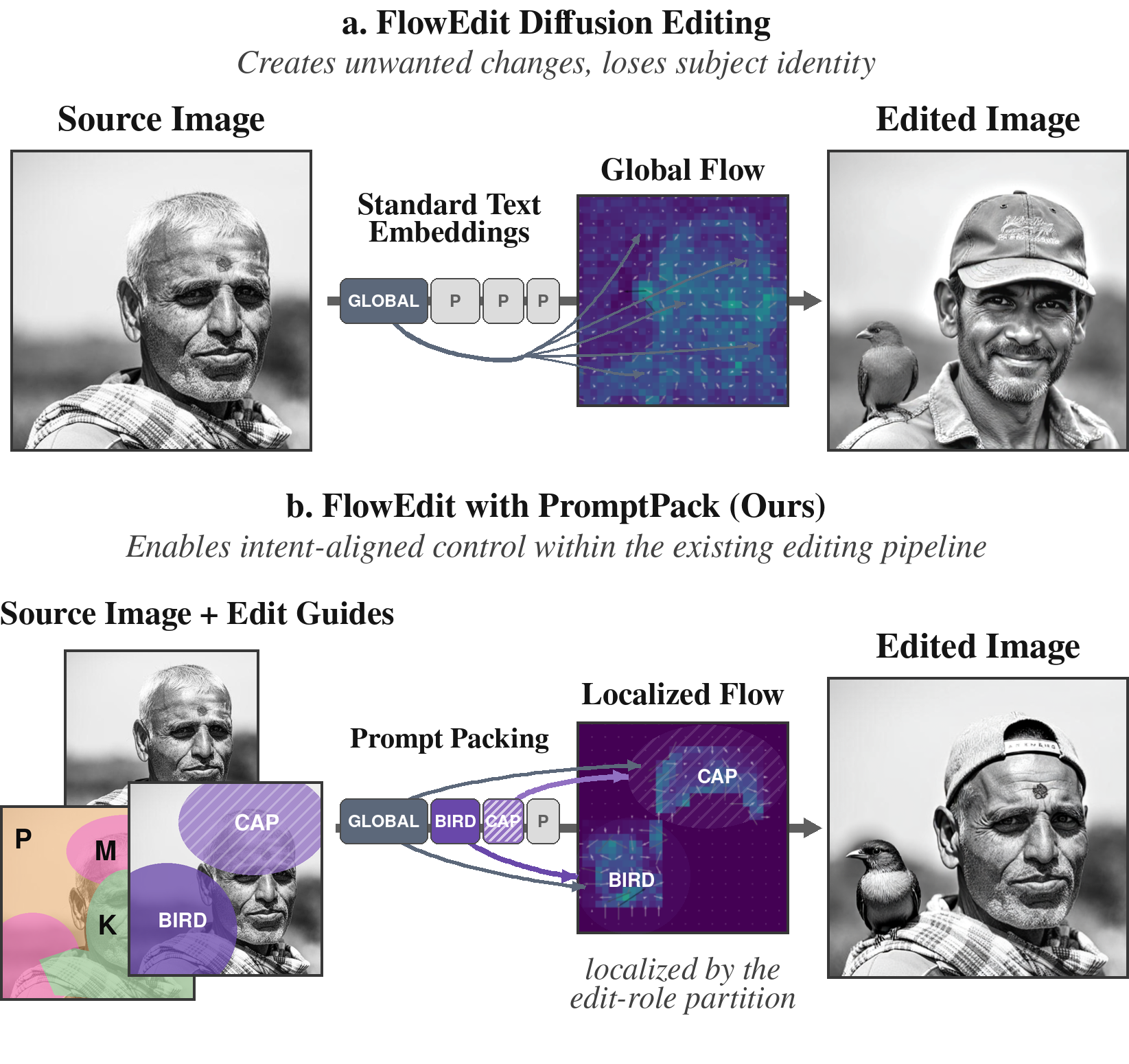}
  \caption{\textbf{PromptPack adds explicit spatial and semantic control to an existing editing pipeline.} (a) FlowEdit: global text conditioning is applied throughout the image, producing unwanted changes and losing subject identity. (b) PromptPack: the user marks where a ``bird'' and a ``cap'' should appear and identifies the man's face and scarf as content to preserve. PromptPack translates these annotations into the edit-role partition and regional-prompt assignment shown at left, packs global and regional text embeddings into the conditioning sequence, and routes them to spatially localized flow supports. This enables intent-aligned control.}
  \label{fig:opener}
\end{figure}

\begin{figure*}[ht]
  \centering
  \makebox[\textwidth]{\includegraphics[width=\textwidth]{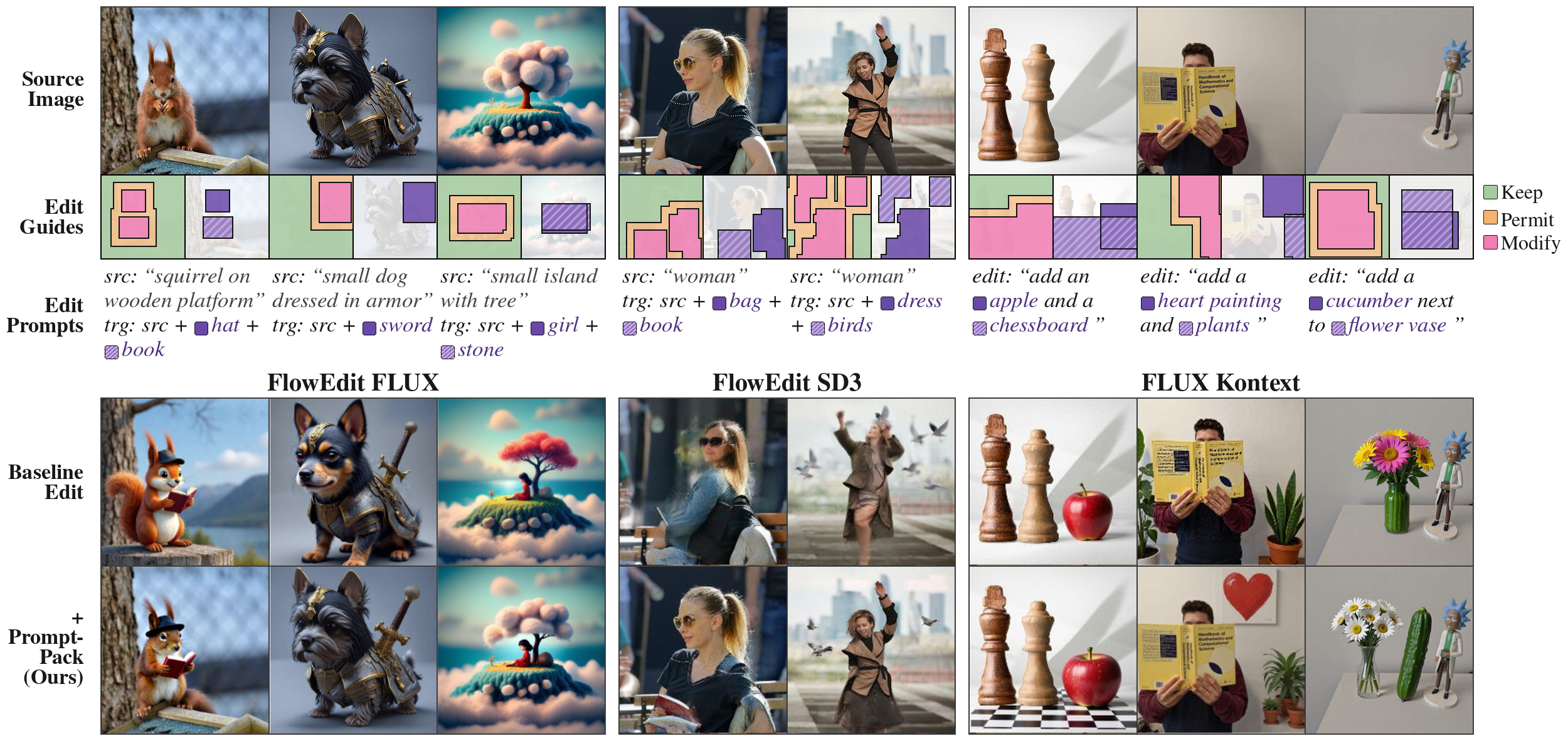}}
  \caption{\textbf{Qualitative comparisons across the three evaluated integrations.} Rows 1 and 3 show the source images and user-provided text required by each baseline editor. FlowEdit FLUX and FlowEdit SD3 use source and target descriptions, whereas FLUX Kontext uses a single edit instruction. Baseline outputs may alter unrelated content or incompletely realize requested concepts (row 4). PromptPack retains the same text conditioning and adds the Edit Guides in row 2, producing outputs that better follow the requested spatial intent while preserving unrelated content (row 5). For these PIE-Bench++ examples, we draw one concept-placement box per requested addition. Each box is an assignment mask; their union initializes \emph{Modify}, a dilation collar defines \emph{Permit}, and the remainder defines \emph{Keep}. Solid and striped swatches link the regional prompts in row 3 to their assignment masks. Architecture-specific mappings are detailed in the supplement.}
  \label{fig:ppflow}
\end{figure*}

Text-guided diffusion editors make complex image manipulation accessible, but a natural-language instruction rarely specifies every pixel that may change. A request such as \emph{``add a scarf to the dog''} leaves several questions unresolved: Must text in the background remain bitwise identical? May the dog's pose shift slightly? Where should the scarf appear? Editors that infer these choices can alter unrelated colors, textures, objects, or identities. Conversely, a strict binary inpainting mask can prevent the local changes needed for a coherent boundary, shadow, or interaction.

We study \emph{interactive, high-intent editing}, where a user identifies important regions in exchange for predictable behavior. We introduce \emph{PromptPack}, an inference-time interface that augments an existing editor with two complementary Edit Guides. The first, an edit-role partition, assigns every pixel to \emph{Modify} (changes are requested), \emph{Permit} (changes are allowed when needed), or \emph{Keep} (changes are prohibited). The second, a regional-prompt assignment, pairs short semantic descriptions with spatial assignment masks. Together, they specify \emph{where} changes may occur and \emph{which} concepts apply to each image region.

Edit Guides can be constructed from user-intent annotations or derived automatically. In the first case, a user marks the image region associated with each requested addition, removal, or modification and separately marks any regions to preserve. Annotations may be boxes or pixel masks and may contain connected or disconnected components. Annotations for different edit concepts may also overlap. By default, each edit-concept annotation becomes its regional-prompt assignment mask, and their union initializes the \emph{Modify} set. Preservation annotations initialize \emph{Keep}, while remaining pixels default to \emph{Permit}. For edits targeting an existing source object, SAM~3~\cite{carion2025sam3segmentconcepts} can instead ground a supplied source-object phrase and automatically construct both guides using Supplementary Eq.~(2).

For each edit concept, PromptPack packs its corresponding text embedding into available positions in the model's fixed-length conditioning sequence. It then routes each embedding to its associated region using its corresponding attention mask, spatially gates flow updates, and then restores source pixels outside of the final edit support. We implement these operations without training or modifying model weights in FlowEdit FLUX, FlowEdit SD3, and FLUX Kontext. Unlike regional prompting that expands conditioning with each region~\cite{tfrp2024} or fixed-length embedding splicing designed for one object~\cite{zhang2024objectadd}, PromptPack retains the editor's original sequence length while routing several independently encoded concepts simultaneously. To our knowledge, it is the first method to multiplex multiple region--prompt pairs within one pretrained diffusion editor's fixed-length text-conditioning sequence and assign each inserted token group its own spatial support.

Our contributions are as follows:
\begin{enumerate}
  \item We identify two complementary sources of ambiguity in localized diffusion editing: \emph{where} changes may occur and \emph{which} prompt concepts apply to each image region. We formalize these controls as Edit Guides comprising an edit-role partition and a regional-prompt assignment.
  \item We introduce fixed-length regional prompt packing: independently encoded regional phrases replace otherwise padded positions in the existing conditioning sequence and are spatially routed without increasing sequence length or modifying model weights.
  \item We couple this conditioning mechanism to a progressive three-state edit gate and protected-pixel restoration in FlowEdit FLUX, FlowEdit SD3, and FLUX Kontext.
  \item We evaluate PromptPack across three editors using measures of regional concept placement and protected-content preservation, compare the three-state edit-role partition with binary edit masks, measure prompt capacity and runtime, and report 2,304 human judgments.
\end{enumerate}

On PIE-Bench++, PromptPack improves regional concept placement and reduces pixel error in protected regions to near zero across all three integrations. Perceptual source retention also improves for both FLUX integrations, and SD3 maintains SSIM while improving regional placement. In a pairwise user study, preference rates favor PromptPack across all three integrations, reaching statistical significance for both FlowEdit FLUX and FLUX Kontext.

\section{Related Work}

\paragraph{Localized and mask-guided editing.}
Binary masks have long been used to localize synthesis in inpainting and diffusion editing~\cite{suvorov2021resolution,lugmayr2022repaintinpaintingusingdenoising,avrahami2023blendedlatent}. DiffEdit~\cite{couairon2023diffedit} estimates an edit mask from differences between source- and target-conditioned predictions. ZONE~\cite{li2024zone} converts an instruction into an edit layer using attention and an off-the-shelf segmenter, while LIME~\cite{simsar2025lime} discovers a region of interest from intermediate features and cross-attention. Given a mask, MAG-Edit~\cite{mao2024magedit} optimizes the latent under spatial attention constraints, KV-Edit~\cite{zhu2025kvedit} preserves background tokens through key--value reuse, SpotEdit~\cite{qin2026spotedit} identifies stable regions and avoids regenerating them, and CannyEdit~\cite{xie2025cannyedit} combines selective structural control with local and global prompt guidance. PromptPack does not claim masks or automated localization as novel. It instead exposes three distinct roles---a core modification, an optional region into which the edit may propagate, and protected content---together with independently routed regional text. BRIDGE~\cite{xiong2026bridge} likewise observes that a coarse binary mask can become an unintended shape prior, but addresses this with a learned two-zone routing module. CARE-Edit~\cite{wang2026careedit} trains condition-specific experts and a router to combine text, masks, and reference images in a contextual editor. PromptPack instead augments frozen editors at inference time and separates edit permission from semantic routing.

\paragraph{Instruction and regional conditioning.}
Instruction-trained editors such as InstructPix2Pix~\cite{brooks2023instructpix2pixlearningfollowimage}, FLUX.1 Kontext~\cite{blackforestlabs2025kontext}, and OmniGen2~\cite{wu2025omnigen2} infer both the editable support and concept grounding from text and the source image. PromptPack complements these models when that inferred spatial intent is insufficiently predictable. ControlNet~\cite{zhang2023addingconditionalcontroltexttoimage}, T2I-Adapter~\cite{mou2023t2iadapterlearningadaptersdig}, and GLIGEN~\cite{Li2023GLIGEN} learn spatial conditioning for generation. At inference time, Prompt-to-Prompt~\cite{prompt2prompt2022}, Attend-and-Excite~\cite{attendexcite2023}, and Training-Free Regional Prompting (TFRP)~\cite{tfrp2024} manipulate attention to improve token--region correspondence; TFRP concatenates independently encoded regional sequences for multi-region FLUX synthesis, increasing conditioning length with the number of regions. ObjectAdd~\cite{zhang2024objectadd} separately encodes one object phrase and splices its content embeddings after the base content within a fixed CLIP sequence, then guides that object toward one box. PromptPack generalizes fixed-length embedding reuse to multiple independent region--prompt pairs: it removes regional padding, inserts each regional content block into otherwise padded positions of the global sequence, and assigns a distinct attention mask to every inserted block. Thus sequence length and global content tokens remain unchanged while several concepts are routed simultaneously. RegionRoute~\cite{chen2026regionroute} trains attention supervision and LoRA experts for single-object regional style transfer with FLUX Kontext. Its Regional Style Editing Score is useful precedent for separately measuring success inside the requested region and preservation outside it, although its trained style-transfer setting differs from multi-concept localized editing. S$^2$Edit~\cite{liu2025s2edit} spatially constrains a learned identity token after per-image fine-tuning, while MagicQuill V2~\cite{liu2026magicquillv2} separates content, spatial, structural, and color intent through trained control branches. PromptPack couples its fixed-length conditioning mechanism to an explicit edit-role partition in frozen editors. Its Edit Guides may be drawn by a user or, for source-object edits, derived from a text-grounded SAM~3 mask~\cite{carion2025sam3segmentconcepts}. The supplement compares these methods by documented capabilities.

\paragraph{Fixed-length text conditioning.}
PromptPack places regional embeddings in positions otherwise occupied by padding; it does not assume those positions are causally empty. Padding Tone~\cite{toker2025paddingtone} shows that padding representations can affect generation during text encoding or diffusion, with behavior depending on model architecture and training. We therefore preserve the conditioning length, explicitly overwrite selected positions, and treat the potential distribution shift and finite token budget as limitations.

\paragraph{Trajectory-guided editing.}
SDEdit~\cite{meng2022sdedit} edits by perturbing an input and denoising under a new condition. FlowEdit~\cite{kulikov2025flowedit} constructs a direct, inversion-free flow between source and target conditions; RF-Edit~\cite{wang2025taming} improves rectified-flow inversion and shares source features, while SplitFlow~\cite{yoon2025splitflow} decomposes and aggregates semantic flow directions. PromptPack builds on FlowEdit: regional prompts alter the target conditioning, while thresholding and spatial gating restrict the estimated difference field. Edit quality therefore remains bounded by the underlying editor, and stronger localization can reduce edit success.

\section{Method}

PromptPack receives two complementary inputs, collectively called Edit Guides: an edit-role partition and a regional-prompt assignment. The regional assignment produces a Packed Prompt and attention mask; the edit-role partition determines flow localization and post-decoding source restoration. The guides may be drawn directly or initialized from a text-grounded source-object mask, as detailed in the supplement.


\subsection{Inputs and Notation}
\label{sec:promptpack-inputs}

Let $X_{\mathrm{src}}$ be a source image, $p_{\mathrm{src}}$ its description, and $p_{\mathrm{global}}$ the target editing prompt.

\paragraph{Edit-role partition.}
The edit-role partition assigns every image pixel to one of three disjoint sets, $\mathcal{K}$ (\emph{Keep}), $\mathcal{P}$ (\emph{Permit}), or $\mathcal{M}$ (\emph{Modify}), which together partition the image domain $\Omega$:
\begin{equation}
  \mathcal{K}\,\dot\cup\,\mathcal{P}\,\dot\cup\,\mathcal{M}=\Omega.
\end{equation}
Pixels in $\mathcal{M}$ are the core of the requested edit; pixels in $\mathcal{P}$ may change to integrate that edit; and pixels in $\mathcal{K}$ must be preserved with pixel-level accuracy.

\paragraph{Regional-prompt assignment.}
We represent the assignment of edit prompts to designated image regions as a set of $R$ pairs $\{(p_i,r_i)\}_{i=1}^{R}$, where $p_i$ is a short regional edit prompt and $r_i\in\{0,1\}^{H\times W}$ is its binary assignment mask. The mask indicates where $p_i$ may influence the image.

\paragraph{Specification and discretization.}
A user's concept-placement annotation initializes both an assignment mask $r_i$ and a component of $\mathcal{M}$; protected annotations become $\mathcal{K}$, and unassigned pixels default to $\mathcal{P}$ (Fig.~\ref{fig:opener}). The two guides nevertheless remain independent. A user may broaden an assignment mask without enlarging the editable support, or specify an asymmetric \emph{Permit} region when content should expand only in a particular direction, such as lengthening a dress downward.
Regions may be boxes or pixel masks and may contain disconnected components. We max-pool the indicator masks of $\mathcal{M}$, $\mathcal{K}$, and $\mathcal{P}$ onto the latent-update grid, yielding the corresponding binary masks $M$, $K$, and $P$. At model block $j$, we separately max-pool each assignment mask $r_i$ onto the image-token grid, yielding $\tilde r_i^j$. Architecture-specific grids and conflict handling are detailed in the supplement.

\subsection{Automatic Edit-Guide Construction}
\label{sec:automatic_edit_guides}

Edit Guides can be drawn by a user or constructed automatically when an edit targets an object already present in $X_{\mathrm{src}}$. Automatic spatial construction requires an aligned text pair $(q_i,p_i)$, where $q_i$ is a source-object query and $p_i$ is its requested replacement, which we supply explicitly.

PromptPack sends $q_i$ to SAM~3~\cite{carion2025sam3segmentconcepts} to locate instances of the source query. Let $S$ be the union of the resulting grounded masks. We set $\mathcal M=S$, derive a symmetric $\mathcal P$ collar around $S$ and the complementary $\mathcal K$, and use a wider dilation of $S$ as the default assignment region $r_i$ for the corresponding edit prompt $p_i$. The wider assignment tolerates boundary uncertainty and token-grid discretization without enlarging the support in which updates are allowed. We provide the exact construction rule and an 11-query grounding screen in the supplementary material.

\begin{figure}[t]
  \centering
  \includegraphics[width=\columnwidth]{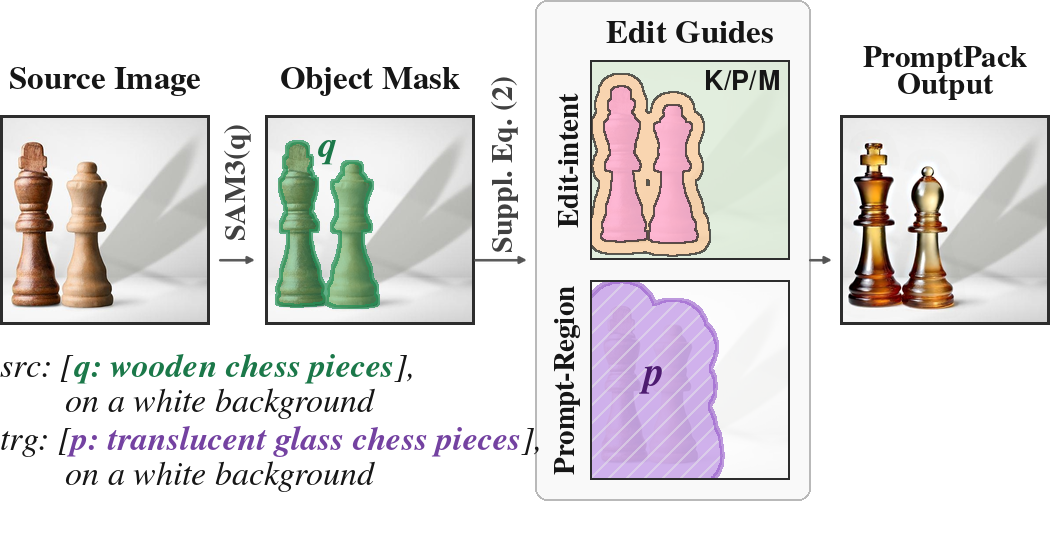}
  \caption{\textbf{Automatic Edit-Guide construction for FlowEdit.} FlowEdit receives the displayed full source and target prompts. The aligned source query $q$ and replacement prompt $p$ are supplied explicitly; automatic construction begins with spatial grounding. SAM~3 receives only $q$ and returns two disconnected masks whose union is $S$. Modify equals $S$; a token-aware collar defines Permit; the remainder defines Keep; and a wider default dilation defines the regional-prompt assignment $r$. PromptPack routes $p$ through $r$ to produce the output.}
  \label{fig:sam3_main}
\end{figure}

\subsection{Prompt Packing}

PromptPack encourages distinct edit concepts to form in specified regions by constructing a \emph{Packed Prompt} and routing its regional tokens during attention redirection (Sec.~\ref{subsection:attention_redirection}). A Packed Prompt inserts independently encoded regional prompt embeddings into padding positions of the global text-conditioning sequence, preserving its fixed length.

Let the base text encoder return $L_{\mathrm{text}}$ vectors of dimension $d_{\mathrm{text}}$. The global prompt occupies $n$ non-padding positions, represented by $E_g\in\mathbb{R}^{n\times d_{\mathrm{text}}}$, and leaves $m=L_{\mathrm{text}}-n$ positions. We encode each regional prompt $p_i$ independently and remove its padding, obtaining $E_i\in\mathbb{R}^{n_i\times d_{\mathrm{text}}}$. We preserve all global content tokens and require $\sum_i n_i\leq m$, a condition satisfied by every prompt in our evaluation. Prompts that exceed this budget must be shortened or omitted, as discussed in Sec.~\ref{sec:limitations}.

Let $E_{\mathrm{pad}}\in\mathbb{R}^{m'\times d_{\mathrm{text}}}$ denote the remaining padding embeddings, where $m'=m-\sum_i n_i$. The packed sequence is
\begin{equation}
  \widetilde E=[E_g;E_1;\ldots;E_R;E_{\mathrm{pad}}]
  \in\mathbb{R}^{L_{\mathrm{text}}\times d_{\mathrm{text}}}.
\end{equation}
Brackets denote concatenation along the token dimension. Because padding embeddings can influence diffusion generation~\cite{toker2025paddingtone}, replacing them may shift the conditioning distribution even though the sequence shape is unchanged.

\paragraph{Editor-specific conditioning.}
FlowEdit uses source and target prompts to estimate a source-to-target difference field. We use the target prompt as $p_{\mathrm{global}}$, pack only the target-conditioned branch, and leave the source-conditioned branch unchanged. FLUX Kontext instead uses a single edit instruction as $p_{\mathrm{global}}$ and augments it with the regional embeddings. Where an editor also uses pooled text conditioning, we retain the pooled embedding of $p_{\mathrm{global}}$; regional prompts enter only through the packed token sequence.

For example, consider changing \emph{``a dog wearing a space suit''} by adding a scarf and flowers in its mouth. FlowEdit receives $p_{\mathrm{src}}=$ \emph{``a dog wearing a space suit''} and $p_{\mathrm{global}}=$ \emph{``a dog wearing a space suit with a scarf and flowers in its mouth.''} FLUX Kontext instead receives the source image and the edit instruction $p_{\mathrm{global}}=$ \emph{``add a scarf and flowers in the dog's mouth.''} Both integrations additionally pack $p_1=$ \emph{``scarf''} and $p_2=$ \emph{``flowers in mouth''} and route them to the neck and mouth regions, respectively.

\begin{figure}[t]
  \centering
  \includegraphics[width=\columnwidth]{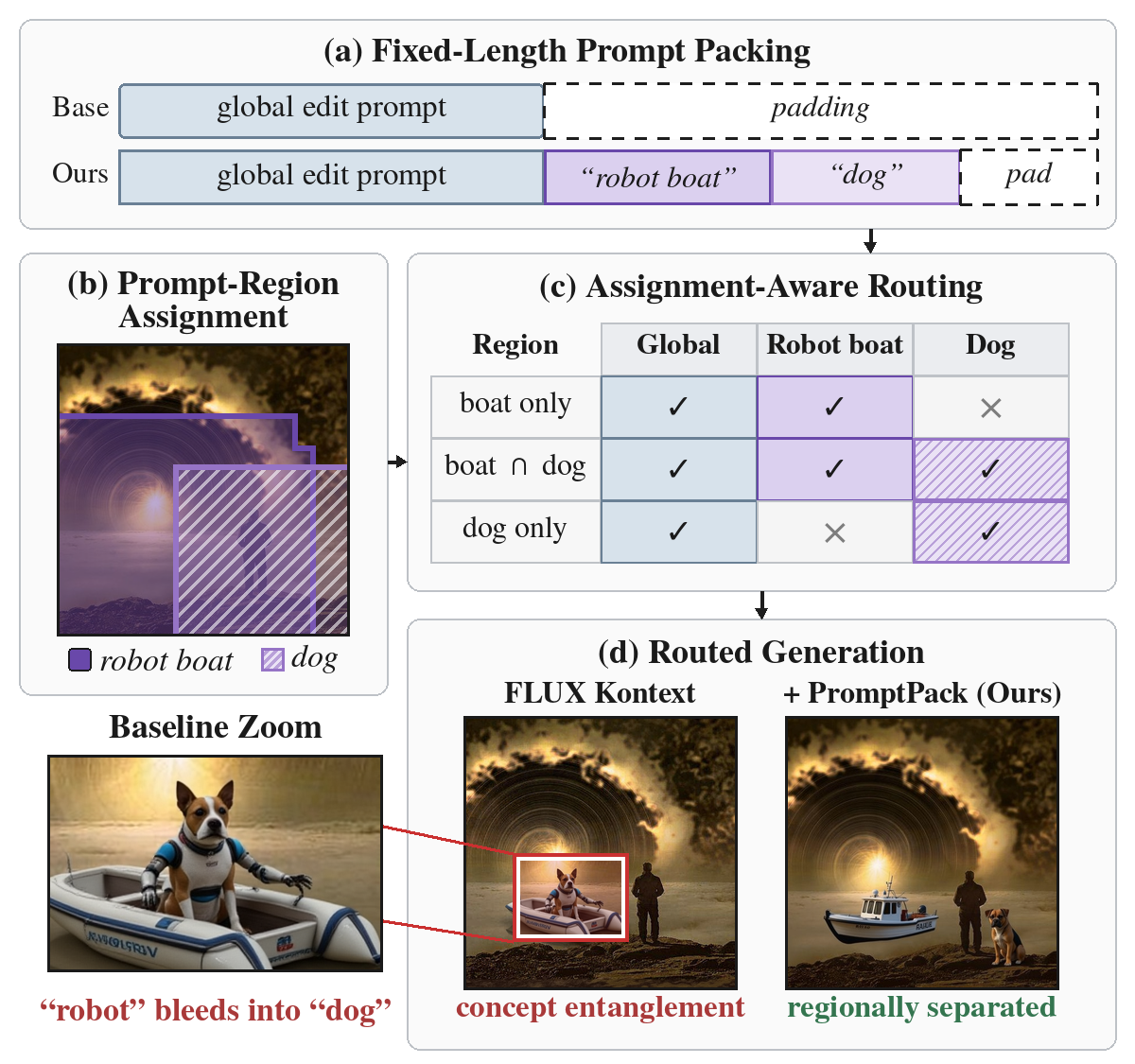}
  \caption{\textbf{Prompt packing and regional routing disentangles concepts.} PromptPack independently encodes regional prompts and packs them into unused conditioning positions (a). Independent embeddings prevent ``robot'' from bleeding into the ``dog'' concept, while spatial routing places the dog to the right of the boat (b--c). The baseline zoom highlights its robot--dog entanglement, whereas PromptPack produces a separate boat and dog in their assigned regions (d).}
  \label{fig:prompt_embeddings}
\end{figure}

\subsection{Attention Redirection}
\label{subsection:attention_redirection}

PromptPack restricts each regional prompt $p_i$ to image patches inside its corresponding region $r_i$ by applying a binary mask during attention computation. At model block $j$, let $L_{\mathrm{img}}^j$ be the number of image tokens and let $\tilde r_i^j\in\{0,1\}^{L_{\mathrm{img}}^j}$ be the flattened assignment mask at that block's spatial resolution. We construct an allowed-connection matrix
\begin{equation}
 A^j=\left[
 \mathbf{1}_{L_{\mathrm{img}}^j\times n},
 \operatorname{rep}(\tilde r_1^j,n_1),\ldots,
 \operatorname{rep}(\tilde r_R^j,n_R),
 \mathbf{1}_{L_{\mathrm{img}}^j\times m'}
 \right],
\end{equation}
where $\operatorname{rep}(\tilde r_i^j,n_i)$ repeats the mask as $n_i$ columns. Thus every image query can use global and remaining padding tokens, whereas a regional token can affect only its assigned image queries.

We convert $A^j$ to an additive logit mask $B^j$ by mapping allowed entries to $0$ and blocked entries to $-\infty$. The redirected attention is
\begin{equation}
 \operatorname{Attn}(Q,K,V)=
 \operatorname{softmax}\!\left(\frac{QK^\top}{\sqrt d}+B^j\right)V.
 \label{eq:masked_logits}
\end{equation}
For cross-attention, $B^j$ masks image-query/text-key connections. For joint-attention models, we apply the mask in both text--image directions while leaving image--image attention unrestricted. Architecture-specific text--text routing is detailed in the supplement.

\begin{figure}[t]
  \centering
  \includegraphics[width=\columnwidth]{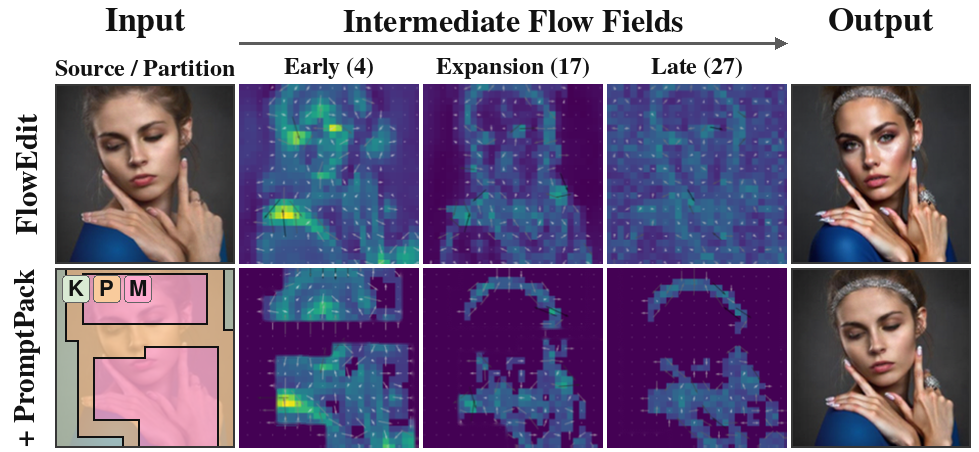}
  \caption{\textbf{Edit-role conditioning localizes intermediate flow.} Both rows add a hair band and diamonds on the woman's hands. FlowEdit receives only the source and applies its difference field globally. PromptPack additionally receives the displayed edit-role partition, restricting early updates to \emph{Modify}, allowing later expansion through \emph{Permit}, and protecting \emph{Keep}. The regional-prompt assignment used by PromptPack is omitted for visual clarity and illustrated separately in Fig.~\ref{fig:prompt_embeddings}.}
  \label{fig:flow_localization}
\end{figure}

\subsection{Edit Localization}
\label{subsection:edit_localization}

To localize image changes according to the specified edit-role partition, PromptPack applies a time-dependent spatial gate to each flow update. Let $s\in\{0,\ldots,T-1\}$ index sampling iterations and let $s_{\mathrm{dil}}$ be the last iteration before the edit may expand. We define
\begin{equation}
 G_s=\begin{cases}
 M,&s\leq s_{\mathrm{dil}},\\
 \operatorname{Dilate}(M,k(s))\cap(M\cup P),&s>s_{\mathrm{dil}},
 \end{cases}
 \label{eq:spatial-gate}
\end{equation}
where $k(s)$ is a monotonically increasing dilation extent. The intersection permits growth from the core edit into $P$ but never into $K$. A narrow \emph{Permit} region behaves like a strict inpainting mask; a wider one gives the generator more freedom to form boundaries, shadows, and local interactions.

\paragraph{PromptPack-Flow (FlowEdit FLUX and SD3).}
FlowEdit~\cite{kulikov2025flowedit} estimates a difference field $V_s^\Delta$ between source- and target-conditioned branches. For each spatial patch $u$, PromptPack applies
\begin{equation}
 \widetilde V_s^\Delta(u)=
 b\,G_s(u)\,\mathbb{1}\!\left[\lVert V_s^\Delta(u)\rVert_2\geq\tau\right]
 V_s^\Delta(u),
 \label{eq:gated-flow}
\end{equation}
then performs the standard FlowEdit update using $\widetilde V_s^\Delta$. The threshold $\tau$ suppresses weak transport signals and $b$ compensates for magnitude lost through thresholding. These terms expose an explicit trade-off: stronger restriction improves preservation but can prevent the target concept from forming.

\paragraph{PromptPack-Generate (FLUX Kontext).}
For a noise-to-image editing trajectory, let $Z_{s+1}^{\mathrm{edit}}$ be the proposed next latent and $Z_{\mathrm{bg},s+1}$ the source-derived background latent at the same noise level. Spatial blending uses the same gate:
\begin{equation}
 Z_{s+1}=G_s\odot Z_{s+1}^{\mathrm{edit}}+(1-G_s)\odot Z_{\mathrm{bg},s+1}.
 \label{eq:generate-blend}
\end{equation}
This formulation retains the edited trajectory inside the gate and the source trajectory outside it.

\subsection{Compression Correction}
\label{sec:compression-correction}

Spatial gating restricts flow updates in latent space, but lossy VAE encoding and decoding can still change protected pixels in the decoded image. PromptPack therefore applies compression correction after decoding, restoring source pixels outside the final edit support. This is particularly useful for preserving source details across multi-turn edits, where reconstruction errors would otherwise accumulate (Supplementary Fig.~\ref{fig:multiturn_full}). Let $U_0$ be the final binary output support on the latent grid and let $\widehat X$ be the decoded edit. We upsample $U_0$ to pixel resolution and optionally smooth its boundary, writing $U=\operatorname{Blur}(\operatorname{Upsample}(U_0);\sigma_{\mathrm{CC}})$, with $\sigma_{\mathrm{CC}}=0$ for a hard support. The delivered image is
\begin{equation}
 X_{\mathrm{out}}=U\odot\widehat X+(1-U)\odot X_{\mathrm{src}}.
 \label{eq:compression-correction}
\end{equation}
With a hard support, Eq.~\ref{eq:compression-correction} exactly restores every pixel outside $U_0$; with boundary smoothing, exact restoration holds outside the transition band. This output constraint explains the near-zero \emph{Keep}-region error.


\section{Experiments}
\label{sec:experiments}

\paragraph{Dataset and Edit Guides.}
We evaluate our method using 80 PIE-Bench++ images containing 160 requested concepts~\cite{UB-CVML-Group_PIE_Bench_pp_2025}, partitioned into a 16-image calibration set and a disjoint 64-image held-out evaluation set. Each image sample is human-annotated to specify the desired edit-concept locations, which we translate into corresponding Edit Guides for evaluation. A separate supplementary screen evaluates Edit Guides automatically derived from text-grounded SAM~3 object masks~\cite{carion2025sam3segmentconcepts}.

\paragraph{Integrations and selection.}
We implement PromptPack in FlowEdit FLUX, FlowEdit SD3, and FLUX Kontext and compare each integration with its base editor using the same source image, prompts, seed 0, sampler, guidance, and all other base-editor settings. FlowEdit FLUX and FLUX Kontext use 28 steps, FlowEdit SD3 uses 50, and dilation begins after iteration 15. We sweep only the FlowEdit threshold $\tau$ and boost $b$ on a deterministic 16-image calibration set (indices $0,5,\ldots,75$), selecting lexicographically by L-ESR, ESR, and SSIM. We then lock $(\tau,b)=(4,1.6)$ for FLUX and $(0.5,1.6)$ for SD3. Table~\ref{tab:main_results} reports only the disjoint 64-image remainder; aggregate metrics over the full 80-image set for the architecture-matched pairs and implementation details are supplementary.

\paragraph{Baselines.}
Our primary comparisons pair each PromptPack integration with its corresponding unmodified base editor. We also evaluate FLUX.1 Fill with binary support, mask-free InstructPix2Pix~\cite{brooks2023instructpix2pixlearningfollowimage}, and three released training-free editors. SpotEdit~\cite{qin2026spotedit} uses its official FLUX-Kontext implementation. For CannyEdit~\cite{xie2025cannyedit}, each original assignment and phrase becomes one binary mask and local prompt; KV-Edit~\cite{zhu2025kvedit} receives the global target and the binary union $M\cup P$. We use released inference settings and do not tune any alternative on our benchmark; exact revisions and settings are supplementary. Note that these methods expose different controls and, except for SpotEdit, use different FLUX editing architectures; their absolute results are therefore provided only as additional comparisons.

\paragraph{Metrics and inference.}
SSIM and LPIPS measure source retention and can reward under-editing. $\mathrm{MSE}_{K}$ measures mean-squared error over the specified Keep set $\mathcal K$ after source restoration and verifies desired edit isolation. CLIP~\cite{radford2021learning} measures global alignment between the edited image and target prompt. We derive three concept-level metrics from YOLO-World v2 detections~\cite{cheng2024yoloworldrealtimeopenvocabularyobject} using a confidence threshold of $0.1$: Edit Success Ratio (ESR), the fraction of requested concepts detected anywhere; Localized Edit Success Ratio (L-ESR), the fraction with a detection centered within the assigned region; and Wrong-Region Rate (WRR), the fraction with a detection centered within another concept's region but outside its own. For each reported paired metric difference (with versus without PromptPack), we estimate a 95\% confidence interval by resampling images with replacement 100,000 times while keeping all concepts from each image together. We repeat this analysis using detector confidence thresholds from $0.1$ to $0.5$ and alternative localization criteria (any overlap and 25/50\% box coverage) for thoroughness.

\paragraph{Capacity and user study.}
Under an IRB-approved protocol, 26 participants completed 2,304 randomized two-alternative judgments (769 FlowEdit FLUX, 768 FLUX Kontext, and 767 FlowEdit SD3; 25 participants saw FlowEdit SD3). We weight participants equally, bootstrap participants as clusters, and apply two-sided Wilcoxon tests against 50\% with Holm correction. Across all evaluated examples, the global and regional prompts fit within each model's fixed-length text-token budget without truncation. Model-specific token usage and remaining capacity are reported in the supplementary material.

\section{Results}

\begin{table*}[t]
\centering
\small
\setlength{\tabcolsep}{4.2pt}
\caption{Held-out evaluation on 64 PIE-Bench++ images and 128 requested concepts after locking the operating points on a disjoint 16-image set. $\mathrm{MSE}_{K}$ verifies the actual Keep-set output constraint. Semantic rates are percentages. Bold compares architecture-matched pairs; released alternatives are contextual.}
\label{tab:main_results}
\begin{tabular*}{\textwidth}{@{\extracolsep{\fill}}l c c c c c c c}
\toprule
\textbf{Method} & \textbf{SSIM}$\uparrow$ & \textbf{LPIPS}$\downarrow$ & $\mathbf{MSE}_{K}\,(10^{-3})\downarrow$ & \textbf{ESR}$\uparrow$ & \textbf{L-ESR}$\uparrow$ & \textbf{WRR}$\downarrow$ & \textbf{CLIP}$\uparrow$ \\
\midrule
FLUX Kontext & .727 & .255 & 8.095 & 57.8 & 32.0 & 14.1 & .320 \\
\textbf{FLUX Kontext + PromptPack} & \textbf{.887} & \textbf{.149} & \textbf{.000} & \textbf{60.2} & \textbf{50.0} & \textbf{13.3} & \textbf{.328} \\
SpotEdit~\cite{qin2026spotedit} & .756 & .235 & 7.722 & 53.1 & 29.7 & 14.8 & .320 \\
\midrule
FlowEdit FLUX & .741 & .280 & 12.710 & 52.3 & 33.6 & 4.7 & .318 \\
\textbf{FlowEdit FLUX + PromptPack} & \textbf{.821} & \textbf{.198} & \textbf{.002} & \textbf{58.6} & \textbf{50.0} & \textbf{1.6} & \textbf{.322} \\
CannyEdit~\cite{xie2025cannyedit} & .600 & .412 & 4.150 & 64.8 & 53.1 & 9.4 & .326 \\
KV-Edit~\cite{zhu2025kvedit} & .767 & .238 & .579 & 49.2 & 32.8 & 9.4 & .315 \\
\midrule
FlowEdit SD3 & .905 & \textbf{.058} & 1.647 & 32.0 & 11.7 & \textbf{5.5} & .301 \\
\textbf{FlowEdit SD3 + PromptPack} & \textbf{.907} & .086 & \textbf{.002} & \textbf{34.4} & \textbf{16.4} & 7.8 & \textbf{.305} \\
\bottomrule
\end{tabular*}
\end{table*}

\paragraph{Preservation and the Keep constraint.}
For FLUX Kontext and FlowEdit FLUX, PromptPack increases SSIM by $+.160$ (95\% CI $[+.110,+.212]$) and $+.080$ ($[+.052,+.110]$), respectively, while reducing LPIPS by $-.106$ and $-.082$. These paired improvements indicate greater structural and perceptual similarity to the source image. Because higher SSIM and lower LPIPS can also result from weaker edits, we interpret these measures alongside the localized concept-adherence results below (L-ESR, ESR, WRR), which show that PromptPack more reliably realizes requested concepts within their assigned regions. For SD3, SSIM is unchanged ($+.002$), while LPIPS increases slightly ($+.027$). Considered alongside the improved localized concept adherence reported below, this pattern is consistent with stronger changes inside the assigned edit regions rather than increased undesired source drift. Keep-set error ($\mathrm{MSE}_{K}$) improves from $1.65$--$12.71\times10^{-3}$ to at most $2.13\times10^{-6}$ in all integrations, verifying that source restoration is properly performed.

\paragraph{Spatial concept adherence.}
PromptPack increases the rate of successful edit localization (L-ESR) by $18.0$ ($[10.1,26.2]$), $16.4$ ($[8.9,24.2]$), and $4.7$ ($[0.8,9.1]$) percentage points for FLUX Kontext, FlowEdit FLUX, and SD3, respectively. PromptPack also raises the overall edit success rate (ESR) by $2.3$ ($[-3.9,8.7]$), $6.3$ ($[0.8,11.9]$), and $2.3$ ($[-2.3,7.0]$) percentage points. The rate at which edit concepts appear in the wrong regions (WRR) decreases by $3.1$ percentage points for FlowEdit FLUX ($[-7.0,0.0]$), but the change is inconclusive for FLUX Kontext and SD3.

\paragraph{Metric sensitivity.}
PromptPack's measured improvement in edit localization (L-ESR) remains consistent across multiple YOLO confidence thresholds and localization criteria. An independent Grounding DINO-Tiny cross-check~\cite{liu2024groundingdino} confirms gains of $8.6$--$14.1$ percentage points, while detector-free regional CLIP~\cite{chen2026regionroute} also yields positive gains for both FLUX integrations. In a method-blinded 30-concept audit, each of three annotators records 7--10 PromptPack wins and no FlowEdit wins. Against human consensus, YOLO L-ESR achieves 95.2\% precision and 63.5\% recall, indicating that it provides a conservative measure of successful localization. Full audit counts, uncertainty estimates, and inter-annotator agreement are provided in the supplementary material.

\paragraph{Comparison with released alternatives.}
PromptPack provides a stronger preservation--localization balance than SpotEdit and KV-Edit. Relative to SpotEdit, FLUX Kontext + PromptPack improves SSIM by $+.131$ and L-ESR by $20.3$ percentage points; relative to KV-Edit, FlowEdit FLUX + PromptPack improves SSIM by $+.054$, L-ESR by $17.2$ percentage points, and reduces WRR by $7.8$ percentage points. CannyEdit attains a higher overall edit-success rate than PromptPack ($64.8\%$ versus $58.6\%$), but PromptPack retains the source substantially more closely (SSIM $.821$ versus $.600$) and produces fewer concepts in incorrect regions (WRR $1.6\%$ versus $9.4\%$). Their localized edit-success rates are similar ($50.0\%$ versus $53.1\%$), with the PromptPack-minus-CannyEdit difference statistically unresolved (95\% CI $[-10.9,+4.7]$ percentage points). Thus, the released comparisons support PromptPack's advantage in jointly preserving source content and controlling concept placement. Note that alternative methods expose different controls and, except for SpotEdit, use different architectures, so our architecture-matched pairs provide the primary evidence of PromptPack's effect. Qualitative comparisons on the same fixed examples across SpotEdit, CannyEdit, KV-Edit, and both architecture-matched base-editor pairs are provided in Supplementary Fig.~\ref{fig:released-alternative-qualitative}.

\paragraph{What do packing and routing contribute?}
PromptPack couples fixed-length packing with spatial attention routing. To separate their contributions, we keep the editor and Edit Guides fixed and compare four conditioning topologies across the held-out evaluation set (Tab.~\ref{tab:packing-topology}). Against the same fixed-length packed embeddings without masks, routing raises L-ESR by $+7.0$ points (95\% CI $[+.8,+13.2]$) and lowers WRR by $-6.3$ ($[-10.6,-2.4]$), with unresolved SSIM change ($-.004$, $[-.011,+.002]$). Thus, routing improves correct concept placement and suppresses cross-region leakage without measurably changing source similarity. Against routed full-length sequences, fixed-length packing has unresolved L-ESR ($+3.1$, $[-4.5,+10.7]$) but lowers WRR by $-4.7$ ($[-8.6,-1.6]$). Its $+4.7$ L-ESR over an ordinary joint prompt is also unresolved, indicating that packing alone does not account for the localization gain. Instead, fixed-length packing retains the spatial-control benefit of routing while reducing peak memory by 10.26 GiB and runtime by 3.27 s/image relative to expanded routing.

\begin{table}[t]
\centering
\scriptsize
\setlength{\tabcolsep}{2.4pt}
\caption{Held-out conditioning-topology diagnostic. $L$ is mean sequence length; time and peak allocation use an RTX PRO 6000.}
\label{tab:packing-topology}
\begin{tabular}{lccccc}
\toprule
\textbf{Conditioning} & $\mathbf{L}$ & \textbf{SSIM}$\uparrow$ & \textbf{L-ESR}$\uparrow$ & \textbf{s}$\downarrow$ & \textbf{GiB}$\downarrow$ \\
\midrule
Ordinary joint & 512 & .836 & 45.3 & 8.21 & 37.55 \\
Packed, no routing & 512 & .824 & 43.0 & 8.13 & 37.74 \\
Expanded routed & 1536 & .813 & 46.9 & 12.96 & 47.99 \\
Packed routed (ours) & 512 & .821 & \textbf{50.0} & 9.70 & 37.74 \\
\bottomrule
\end{tabular}
\end{table}

\paragraph{Overhead.}
On ten fixed held-out images after warm-up, FlowEdit FLUX averages $7.95\pm0.17$ s/image with 37.55 GiB peak memory allocation, compared with $9.34\pm0.17$ s/image and 37.74 GiB for FlowEdit + PromptPack on an RTX PRO 6000. Accordingly, the complete PromptPack path, including packing, masks, and correction, incurs a runtime overhead of 1.40 s/image (17.6\%) and a peak-memory overhead of 0.19 GiB (0.5\%).

\paragraph{Human preference and global alignment.}
Our 26-participant user study finds that the mean participant-level preference for PromptPack over baseline FlowEdit FLUX is 87.2\% (95\% CI $[83.9,90.4]$; Holm $p<.001$). The user-preference rate is also 67.7\% over baseline FLUX Kontext ($[60.5,73.5]$; $p<.001$) and 54.7\% over FlowEdit SD3 ($[50.6,59.0]$; $p=.055$). CLIP scores differ by less than $.01$ for every final method pair, indicating that PromptPack improves spatial behavior while preserving comparable global prompt alignment.

\paragraph{Does a three-state edit-role gate improve localization?}
We compare progressive K/P/M gating with a tight binary gate restricted to $M$ and a wide gate active throughout $M\cup P$ (Tab.~\ref{tab:main-kpm-screen}; see visual results in Supplementary Fig.~\ref{fig:kpm-binary}). On the first 20 held-out images under two seeds, K/P/M improves L-ESR over the tight gate by 2.5 points (95\% CI $[0,6.3]$) with a small SSIM reduction of $-.009$ ($[-.012,-.006]$). Relative to the wide gate, it improves SSIM by $.076$ ($[.056,.097]$) and reduces raw Keep error by $.469\times10^{-3}$, while the L-ESR difference remains unresolved. Progressive gating therefore balances the tight gate's lower edit success against the wide gate's source drift. K/P/M also exceeds matched FLUX inpainting by $.202$ SSIM and 13.8 L-ESR points. In a separate directional screen, asymmetric \emph{Permit} regions reduce editable support without lowering observed L-ESR.

\begin{table}[t]
\centering
\small
\setlength{\tabcolsep}{4pt}
\caption{Fixed held-out K/P/M screen. Raw $\mathrm{MSE}_{K}$ is before source restoration; detection rates are percentages.}
\label{tab:main-kpm-screen}
\begin{tabular}{lccc}
\toprule
\textbf{Gate / editor} & \textbf{SSIM}$\uparrow$ & $\mathbf{MSE}^{raw}_{K}(10^{-3})\downarrow$ & \textbf{L-ESR}$\uparrow$ \\
\midrule
Tight ($M$) & .858 & .658 & 40.0 \\
Wide ($M\cup P$) & .773 & 1.135 & 45.0 \\
K/P/M & .849 & .666 & 42.5 \\
FLUX inpaint & .647 & 1.116 & 28.7 \\
\bottomrule
\end{tabular}
\end{table}

\section{Conclusion}

PromptPack combines fixed-length regional prompt packing with Edit Guides that separate semantic assignment from edit permission. Across three frozen diffusion editors, it improves localized concept placement while enforcing specified Keep regions, providing spatial control for both source--target and instruction-driven editing without retraining.

\paragraph{Limitations.}
\label{sec:limitations}
Automatic grounding applies only to existing source objects. Packing has finite capacity and may perturb nonempty padding representations~\cite{toker2025paddingtone}. Integrations require architecture-specific attention and latent access; source restoration may introduce seams, and outputs remain bounded by the base editor.

{
\small
\bibliographystyle{ieeenat_fullname}
\bibliography{main}
}
\clearpage
\phantomsection
\addcontentsline{toc}{section}{Supplementary Material}
\begin{center}
{\Large\bfseries Supplementary Material}
\end{center}
\section*{Contents}
\setcounter{section}{0}
\renewcommand{\thesection}{S\arabic{section}}
\renewcommand{\theHsection}{supp.\arabic{section}}
\hypersetup{allcolors=black}
\begingroup
\large
\raggedright
\newcommand{\suppcontentsentry}[2]{%
  \hyperref[#2]{#1}\nobreak\dotfill\nobreak\hyperref[#2]{\pageref*{#2}}}
\newcommand{\suppsubentry}[3]{\hyperref[#3]{\textbf{#1}\enspace #2}}
\begin{enumerate}[label=S\arabic*.]\setlength{\itemsep}{0.6em}
    \item \suppcontentsentry{\textbf{PromptPack Integration}}{supp:integration-map}
  \begin{itemize}[label={},leftmargin=1.6em]
    \item \suppsubentry{S1.1}{Full implementation algorithms (PromptPack-Flow and PromptPack-Generate)}{supp:promptpack-formulations}
    \item \suppsubentry{S1.2}{Implementation details}{supp:spatial-routing}
    \begin{itemize}[label={},leftmargin=1.4em,itemsep=0.1em,topsep=0.1em]
      \item \suppsubentry{S1.2.1}{Attention redirection}{supp:attention-redirection-implementation}
      \item \suppsubentry{S1.2.2}{Spatial gating}{supp:spatial-gating-implementation}
      \item \suppsubentry{S1.2.3}{Compression correction}{supp:compression-correction}
    \end{itemize}
    \item \suppsubentry{S1.3}{Hyperparameters and spatial gate dilation schedules}{supp:hyperparameters}
    \item \suppsubentry{S1.4}{Prompt capacity and computational overhead}{supp:prompt-capacity}
  \end{itemize}

    \item \suppcontentsentry{\textbf{Automatic Edit-Guide Construction}}{sec:sam3_automatic_grounding}
  \begin{itemize}[label={},leftmargin=1.6em]
    \item \suppsubentry{S2.1}{SAM~3 grounding and Edit-Guide derivation}{supp:sam3-grounding-derivation}
    \item \suppsubentry{S2.2}{Identifying regional dependencies}{supp:regional-dependencies}
    \item \suppsubentry{S2.3}{Automatic-grounding evaluation}{supp:automatic-grounding-evaluation}
  \end{itemize}

    \item \suppcontentsentry{\textbf{Evaluation Protocol}}{supp:evaluation-protocol}
  \begin{itemize}[label={},leftmargin=1.6em]
    \item \suppsubentry{S3.1}{Dataset split and Edit-Guide annotations}{supp:dataset-edit-guide-annotations}
    \item \suppsubentry{S3.2}{Metric definitions and statistical inference}{supp:metric-definitions}
    \item \suppsubentry{S3.3}{User-study protocol}{supp:user-study}
  \end{itemize}

    \item \suppcontentsentry{\textbf{Additional Quantitative Evaluation and Robustness}}{supp:complete-results}
  \begin{itemize}[label={},leftmargin=1.6em]
    \item \suppsubentry{S4.1}{Complete benchmark results}{supp:complete-benchmark-results}
    \item \suppsubentry{S4.2}{Held-out paired uncertainty}{supp:heldout-paired-uncertainty}
    \item \suppsubentry{S4.3}{Quantitative comparisons with released alternatives}{supp:released-alternatives}
    \item \suppsubentry{S4.4}{Spatial-metric sensitivity}{supp:spatial-metric-sensitivity}
    \item \suppsubentry{S4.5}{Capability comparison and scope}{supp:capability-scope}
  \end{itemize}

\newpage
\vspace*{0.20em}
    \item \suppcontentsentry{\textbf{Additional Qualitative Results}}{supp:multi-turn}
  \begin{itemize}[label={},leftmargin=1.6em]
    \item \suppsubentry{S5.1}{PIE-Bench++ visual comparisons}{supp:piebench-visual-comparisons}
    \item \suppsubentry{S5.2}{Component and flow-localization ablations}{supp:qualitative-ablations}
    \item \suppsubentry{S5.3}{Failure cases}{supp:failure-cases}
  \end{itemize}

\begin{samepage}
    \vspace*{0.45em}
    \item \suppcontentsentry{\textbf{Design Choice Ablations}}{supp:design-ablations}
  \begin{itemize}[label={},leftmargin=1.6em]
    \item \suppsubentry{S6.1}{Fixed-length prompt packing and regional routing}{supp:packing-routing}
    \item \suppsubentry{S6.2}{Prompt-capacity stress test}{supp:capacity-stress}
    \item \suppsubentry{S6.3}{Three-state spatial gating versus binary masks}{supp:three-state-gating}
    \item \suppsubentry{S6.4}{Asymmetric Permit regions}{supp:asymmetric-permit}
    \item \suppsubentry{S6.5}{Flow-magnitude threshold}{supp:flow-threshold}
    \item \suppsubentry{S6.6}{Flow-boost factor}{supp:flow-boost}
    \item \suppsubentry{S6.7}{Component ablations}{supp:component-ablations}
  \end{itemize}
\end{samepage}

    \item \suppcontentsentry{\textbf{Exploratory Video Extension}}{supp:video-extension}
\end{enumerate}
\endgroup
\hypersetup{allcolors=wacvblue}
\clearpage

\section{PromptPack Integration}
\label{supp:integration-map}
\begin{figure*}[t]
    \centering
      \includegraphics[width=0.98\textwidth]{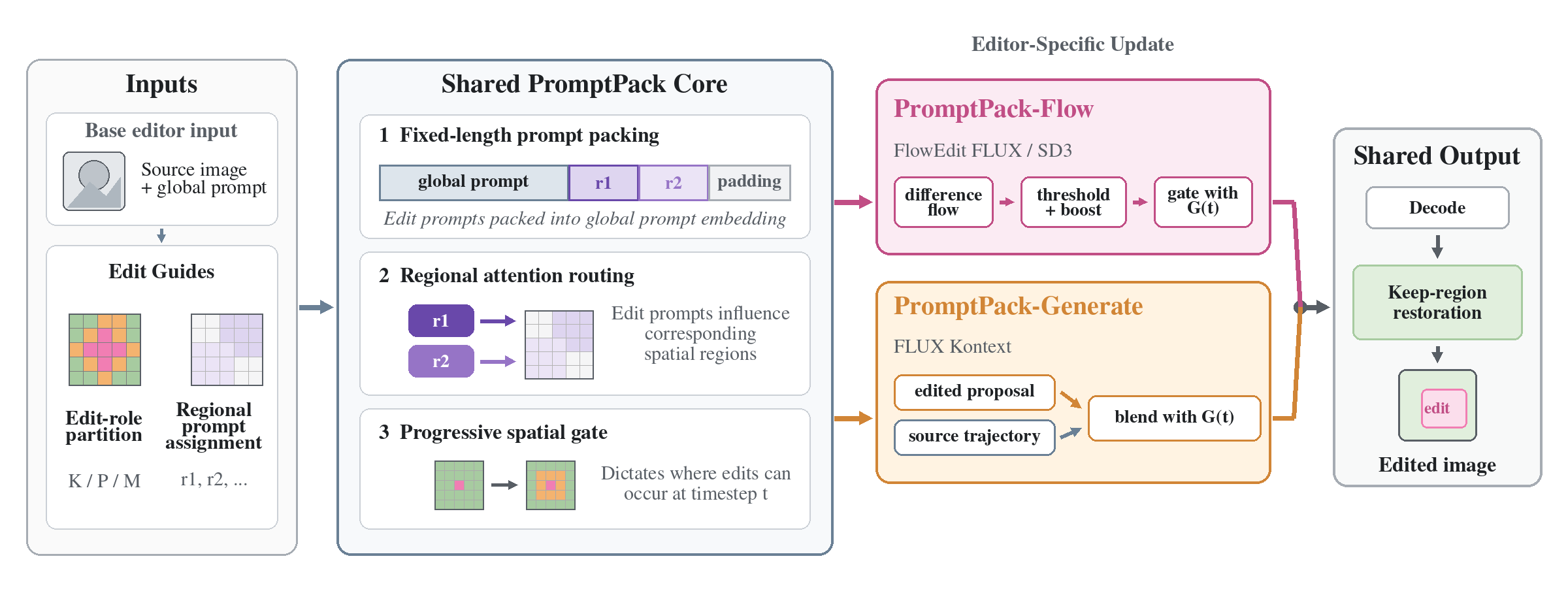}
  \caption{\textbf{Shared PromptPack core and editor-specific integration points.} PromptPack packs independently encoded regional text into the editor's fixed-length conditioning sequence, routes each group with its assignment mask, and derives a progressive gate from the edit-role partition. PromptPack-Flow applies this gate to a thresholded and boosted source-to-target difference field; PromptPack-Generate blends edited and source-derived latent trajectories. Both formulations decode the result and restore protected source pixels.}
  \label{fig:system_diagram}
\end{figure*}

\subsection{Full implementation algorithms (PromptPack-Flow and PromptPack-Generate)}
\label{supp:promptpack-formulations}
All three integrations of PromptPack (FlowEdit FLUX, FlowEdit SD3, and FLUX
Kontext) use the same fixed-length prompt packing, regional attention routing,
and progressive spatial gating (Fig.~\ref{fig:system_diagram}). FlowEdit-based
editors (FlowEdit FLUX and FlowEdit SD3) use PromptPack-Flow, which gates
FlowEdit's source-to-target difference field. Direct noise-to-image editors
(FLUX Kontext) use PromptPack-Generate, which blends the edited trajectory with
a source-derived background trajectory at each sampling step.

Full implementation Algorithms~\ref{alg:promptpackflow} and~\ref{alg:promptpackgenerate} use the same notation and gate $G_s$ as the main paper. Architecture-specific routines implement $\operatorname{Pack}$ and $\operatorname{RedirectedVelocity}$ by replacing prompt embeddings and installing the additive attention masks described in Eq.~\ref{eq:masked_logits}.

\subsection{Implementation details}
\label{supp:spatial-routing}
PromptPack converts the two pixel-space Edit Guides into the attention masks
and spatial gates consumed by Algorithms~\ref{alg:promptpackflow}
and~\ref{alg:promptpackgenerate}, as described below.

\subsubsection{Attention redirection}
\label{supp:attention-redirection-implementation}
Assignment masks are max-pooled onto each editor's image-token grid; a grid cell
is active when any source pixel mapped to it is active. For image-query/text-key
cross-attention, the additive mask in Eq.~\ref{eq:masked_logits} is applied
directly. For joint-attention models such as FLUX, we place the
allowed-connection matrix and its transpose in the two off-diagonal text--image
blocks while leaving image--image attention unrestricted. FLUX retains
unrestricted text--text attention, whereas the SD3 and FLUX Kontext integrations
use prompt-wise block-diagonal text--text attention. Both FlowEdit variants
route assignments at $32\!\times\!32$, while FLUX Kontext routes at
$64\!\times\!64$ and broadens assignments through sequential width-30 and
width-10 max-pooling.

\subsubsection{Spatial gating}
\label{supp:spatial-gating-implementation}
Edit-role masks are max-pooled onto each editor's latent-update grid to construct
$G_s$; a grid cell is active when any source pixel mapped to it is active. If
multiple roles map to one cell, we use the priority \emph{Modify}, then
\emph{Keep}, then \emph{Permit}. FlowEdit FLUX gates updates at
$32\!\times\!32$; FlowEdit SD3 and FLUX Kontext gate at $64\!\times\!64$.

\newpage
\begin{algorithm}[H]
\caption{PromptPack-Flow (FlowEdit)}
\label{alg:promptpackflow}
\begin{algorithmic}[1]
\Require Source image $X_{\mathrm{src}}$ and latent $Z_{\mathrm{src}}$, source prompt $p_{\mathrm{src}}$, target prompt $p_{\mathrm{global}}$, regional pairs $\{(p_i,r_i)\}_{i=1}^R$, intent masks $K,P,M$, descending time grid $t_0>\cdots>t_T=0$, threshold $\tau$, boost $b$
\Ensure Edited image $X_{\mathrm{out}}$
\State $\widetilde E,B\gets\operatorname{PackAndMask}(p_{\mathrm{global}},\{(p_i,r_i)\})$
\State $Z^{\mathrm{edit}}_0\gets Z_{\mathrm{src}}$
\For{$s=0$ to $T-1$}
  \State $N_s\sim\mathcal{N}(0,I)$
  \State $Z^{\mathrm{src}}_s\gets(1-t_s)Z_{\mathrm{src}}+t_sN_s$
  \State $Z^{\mathrm{tar}}_s\gets Z^{\mathrm{edit}}_s+Z^{\mathrm{src}}_s-Z_{\mathrm{src}}$
  \State $V^{\mathrm{tar}}_s\gets v_\theta(Z^{\mathrm{tar}}_s,t_s,\widetilde E;B)$
  \State $V^{\mathrm{src}}_s\gets v_\theta(Z^{\mathrm{src}}_s,t_s,p_{\mathrm{src}})$
  \State $V_s^\Delta\gets V^{\mathrm{tar}}_s-V^{\mathrm{src}}_s$
  \State $\widetilde V_s^\Delta\gets bG_s\odot\mathbb{1}[\lVert V_s^\Delta\rVert_2\geq\tau]\odot V_s^\Delta$
  \State $Z^{\mathrm{edit}}_{s+1}\gets Z^{\mathrm{edit}}_s+(t_{s+1}-t_s)\widetilde V_s^\Delta$
\EndFor
\State $\widehat X\gets\operatorname{Decode}(Z^{\mathrm{edit}}_T)$
\State $U_0\gets\operatorname{OutputSupport}(M)$
\State $X_{\mathrm{out}}\gets\operatorname{CompressionCorrection}(X_{\mathrm{src}},\widehat X,U_0)$
\State \Return $X_{\mathrm{out}}$
\end{algorithmic}
\end{algorithm}

\begin{algorithm}[H]
\caption{PromptPack-Generate (FLUX Kontext)}
\label{alg:promptpackgenerate}
\begin{algorithmic}[1]
\Require Source image $X_{\mathrm{src}}$ with chronological source-derived trajectory $\{Z_{\mathrm{bg},s}\}_{s=0}^T$, target prompt $p_{\mathrm{global}}$, regional pairs $\{(p_i,r_i)\}_{i=1}^R$, intent masks $K,P,M$, time grid $t_0,\ldots,t_T$
\Ensure Edited image $X_{\mathrm{out}}$
\State $\widetilde E,B\gets\operatorname{PackAndMask}(p_{\mathrm{global}},\{(p_i,r_i)\})$
\State $Z_0\sim\mathcal{N}(0,I)$
\For{$s=0$ to $T-1$}
  \State $V_s\gets v_\theta(Z_s,t_s,\widetilde E;B)$
  \State $Z_{s+1}^{\mathrm{edit}}\gets\operatorname{Sample}(Z_s,V_s,t_s,t_{s+1})$
  \State $Z_{s+1}\gets G_s\odot Z_{s+1}^{\mathrm{edit}}+(1-G_s)\odot Z_{\mathrm{bg},s+1}$
\EndFor
\State $\widehat X\gets\operatorname{Decode}(Z_T)$
\State $U_0\gets\operatorname{OutputSupport}(M)$
\State $X_{\mathrm{out}}\gets\operatorname{CompressionCorrection}(X_{\mathrm{src}},\widehat X,U_0)$
\State \Return $X_{\mathrm{out}}$
\end{algorithmic}
\end{algorithm}

\newpage
\vspace*{-0.05cm}
\subsubsection{Compression correction}
\label{supp:compression-correction}

To correct for pixel perturbations introduced by an editor's VAE, PromptPack
combines the decoded edit $\widehat X$ with the source image $X_{\mathrm{src}}$
after sampling. It derives a final binary output support
$U_0=\operatorname{OutputSupport}(M)$ from the requested Modify region, upsamples
that support to pixel resolution, and optionally smooths its boundary. Pixels
outside a hard support are copied directly from the source; smoothing instead
creates a narrow transition band. Algorithm~\ref{alg:compression_correction}
gives the complete image-space operation.

\begin{algorithm}[H]
\caption{CompressionCorrection}
\label{alg:compression_correction}
\begin{algorithmic}[1]
\Require Source image $X_{\mathrm{src}}$, decoded edit $\widehat X$, binary output support $U_0$, optional boundary blur $\sigma_{\mathrm{CC}}$
\Ensure Corrected image $X_{\mathrm{out}}$
\State $U\gets\operatorname{Upsample}(U_0)$
\If{$\sigma_{\mathrm{CC}}>0$}
  \State $U\gets\operatorname{GaussianBlur}(U,\sigma_{\mathrm{CC}})$
\EndIf
\State $X_{\mathrm{out}}\gets U\odot\widehat X+(1-U)\odot X_{\mathrm{src}}$
\State \Return $X_{\mathrm{out}}$
\end{algorithmic}
\end{algorithm}

\paragraph{Preservation guarantee and $\mathrm{MSE}_{K}$.}
Because the spatial gate $G_s\subseteq M\cup P$ for every $s$, the Keep set $K$ lies outside the sampling support. With a hard output support, Eq.~\ref{eq:compression-correction} gives $X_{\mathrm{out}}(u)=X_{\mathrm{src}}(u)$ wherever $U_0(u)=0$. With boundary smoothing, exact equality applies only outside the soft transition band. We evaluate $\mathrm{MSE}_{K}$ over all of $K$; its small nonzero value includes any transition-band overlap and therefore verifies the delivered-image constraint rather than generative quality.

\subsection{Hyperparameters and spatial gate dilation schedules}
\label{supp:hyperparameters}
\subsubsection{Hyperparameters}
Table~\ref{tab:promptpack_hparams} records the settings used for the quantitative evaluation. A dash marks an inapplicable operation.
\begin{table}[h!]
  \centering
  \scriptsize
  \setlength{\tabcolsep}{2.5pt}
  \renewcommand{\arraystretch}{1.1}
  \begin{tabular}{l c c c}
    \toprule
    \textbf{Setting} & \textbf{FlowEdit FLUX} & \textbf{FlowEdit SD3} & \shortstack{\textbf{FLUX}\\\textbf{Kontext}} \\
    \midrule
    Sampling steps & 28 & 50 & 28 \\
    Dilation begins after & 15 & 15 & 15 \\
    Magnitude threshold $\tau$ & 4 & 0.5 & -- \\
    Flow boost $b$ & 1.6 & 1.6 & -- \\
    Output-support kernel & 10 & 10 & 10 \\
    Boundary blur $\sigma_{\mathrm{CC}}$ & 10 & 10 & 10 \\
    \bottomrule
  \end{tabular}
  \caption{Quantitative-evaluation settings. Output-support kernels act on the $64\!\times\!64$ mask before pixel-space upsampling.}
  \label{tab:promptpack_hparams}
\end{table}

In FlowEdit, $\tau$ removes weak difference-field updates, flow boost $b$
scales the retained updates, and boundary blur $\sigma_{\mathrm{CC}}$ controls
the final output-boundary blend.

\paragraph{Locked operating-point selection.}
We select the FlowEdit FLUX and FlowEdit SD3 operating points on a deterministic 16-image PIE-Bench++ calibration subset (indices $0,5,\ldots,75$). The calibration sweep uses seed 0 and the same final-output YOLO-World protocol as the main evaluation; sampling, guidance, and all parameters other than $\tau$ and $b$ remain fixed. Among those eligible settings, we select lexicographically by L-ESR, ESR, and then SSIM. This yields $(\tau,b)=(4,1.6)$ for FlowEdit FLUX and $(0.5,1.6)$ for FlowEdit SD3. We then freeze these values before evaluating the remaining 64 images.

\subsubsection{Spatial gate dilation schedule}
For FlowEdit FLUX and SD3, PromptPack uses a time-varying spatial gate $G_s$ to
determine where edits may occur at sampling iteration $s$. The gate equals the
Modify set $M$ through iteration 15; thereafter, at iteration $s$, its
max-pooling kernel width is
\begin{equation}
k_{\mathrm{flow}}(s)=3+\left\lfloor\frac{s-15}{3}\right\rfloor,\qquad s>15.
\end{equation}
For 28-step FLUX sampling this progresses from width 3 at iteration 16 to width 7 at iteration 27. SD3 follows the same rule over 50 iterations. The user-facing Permit set is the additional support reachable under this schedule; equivalently, intersecting with $M\cup P$ leaves the implemented gate unchanged while identifying the rest as Keep.

The FLUX Kontext integration uses repeated $3\!\times\!3$ dilation, with $1+2(s-15)$ iterations for $s\geq15$, and applies a gate blur of $\sigma_g=2.5$ from iteration 20 onward. This more aggressive schedule belongs to PromptPack-Generate, where the edited trajectory begins farther from the source trajectory. Compression correction is separate from both schedules and uses the output-support settings in Tab.~\ref{tab:promptpack_hparams}.

\begin{table}[h!]
  \centering
  \scriptsize
  \setlength{\tabcolsep}{2.5pt}
  \renewcommand{\arraystretch}{1.1}
  \begin{tabular}{l c c c}
    \toprule
    \textbf{Integration} & \textbf{Grid} & \textbf{Dilation parameter} & \textbf{Late blur} \\
    \midrule
    FlowEdit FLUX & $32\!\times\!32$ & max-pool width $k_{\mathrm{flow}}$ & none \\
    FlowEdit SD3 & $64\!\times\!64$ & max-pool width $k_{\mathrm{flow}}$ & none \\
    FLUX Kontext & $64\!\times\!64$ & repeated $3\!\times\!3$ dilation & $\sigma_g=2.5$ \\
    \bottomrule
  \end{tabular}
  \caption{Architecture-specific gate implementations. The formula in the text gives the iteration schedule.}
  \label{tab:promptpack_flow_dilation}
\end{table}

\subsection{Prompt capacity and computational overhead}
\label{supp:prompt-capacity}
Fixed-length packing raises two practical questions: whether the global and
regional prompts fit without truncation, and what inference cost is added by
packing, routing, and compression correction. To answer the first, we reproduce
each integration's exact normalization and tokenize all 80 examples on CPU.
Table~\ref{tab:capacity} reports regional-token usage, total occupied sequence
positions, and remaining capacity; every evaluated prompt fits without
truncation. The following runtime study answers the second question.

\begin{table}[t]
\centering
\small
\setlength{\tabcolsep}{3.5pt}
\caption{Packed-token usage. Values are median (maximum or minimum).}
\label{tab:capacity}
\begin{tabular}{lccc}
\toprule
\textbf{Integration} & \textbf{Regional added} & \textbf{Total used} & \textbf{Remaining} \\
\midrule
FlowEdit FLUX & 9 (19) & 32 (52) / 512 & 480 (460) \\
FLUX Kontext & 9 (19) & 18 (37) / 512 & 494 (475) \\
FlowEdit SD3 & 10 (19) & 30 (48) / 333 & 303 (285) \\
\bottomrule
\end{tabular}
\end{table}

\subsubsection{Runtime and memory overhead}
We benchmark the final FlowEdit FLUX operating point on ten fixed held-out
images after one untimed warm-up image.
Each method uses seed 0 and 28 sampling steps on the same RTX PRO 6000. Timing
includes prompt and mask preparation, VAE encode/decode, editing, and, for
PromptPack, compression correction. CUDA peaks are reset immediately before
each run. Table~\ref{tab:overhead} reports means across images; runtime
variation is the population standard deviation.

\begin{table}[t]
\centering
\small
\setlength{\tabcolsep}{5pt}
\caption{Matched end-to-end inference overhead.}
\label{tab:overhead}
\begin{tabular}{lcc}
\toprule
\textbf{Method} & \textbf{Time (s)} & \textbf{Peak alloc. (GiB)} \\
\midrule
FlowEdit FLUX & $7.95\pm0.17$ & 37.55 \\
+ PromptPack & $9.34\pm0.17$ & 37.74 \\
\midrule
Overhead & +17.6\% & +0.5\% \\
\bottomrule
\end{tabular}
\end{table}

\clearpage

\section{Automatic Edit-Guide Construction}
\label{sec:sam3_automatic_grounding}
This section asks whether source-object edits can obtain PromptPack's spatial
controls without drawing masks manually. We ground a supplied source-object
phrase with SAM~3, convert the resulting mask into the three edit roles and a
regional-prompt assignment, optionally identify dependent regions such as a
reflection, and then evaluate the resulting guides in a fixed one-shot screen.

\subsection{SAM~3 grounding and Edit-Guide derivation}
\label{supp:sam3-grounding-derivation}
We test whether PromptPack's spatial inputs can be constructed without manual annotation when an edit targets an object already present in the source image. Each case supplies an aligned source-query/replacement pair $(q_i,p_i)$ in addition to the full source and target prompts; this screen evaluates automatic spatial grounding rather than automatic prompt-span parsing. Given $q_i$, SAM~3~\cite{carion2025sam3segmentconcepts} predicts masks, confidence scores, and dense foreground probabilities. We select the highest-scoring mask above $0.50$ for a singular query and union all above-threshold instances for a plural query. The requested replacement $p_i$ remains the independently supplied regional prompt; SAM~3 grounds $q_i$ spatially but does not generate or receive $p_i$.

\paragraph{Edit Guide construction.}
We use one deterministic rule to construct the edit-role partition and regional
prompt-assignment mask from the selected SAM~3 mask. At $512\!\times\!512$, the FLUX transformer operates on a $32\!\times\!32$ spatial grid, so one token spans $\ell=16$ image pixels. Let $S$ denote the selected SAM~3 mask. Let $F_{.25}$ be the connected component containing $S$ in the SAM~3 foreground support thresholded at probability $0.25$, and let
\begin{equation}
d_{95}=Q_{.95}\{\operatorname{dist}(u,S):u\in F_{.25}\setminus S\}.
\end{equation}
We set $r_P=\operatorname{clip}(\operatorname{round}(d_{95}),2\ell,3\ell)$ and $r_R=\min(r_P+2\ell,5\ell)$. The resulting guides are
\begin{equation}
\begin{aligned}
\mathcal M&=S, & \mathcal P&=D(S,r_P)\setminus S,\\
\mathcal K&=\Omega\setminus D(S,r_P), & r_i&=D(S,r_R).
\end{aligned}
\label{eq:sam3-controls}
\end{equation}
If the uncertain exterior is empty, clipping selects the two-token lower bound. At this resolution, the rule yields a 32--48 pixel Permit collar and a 64--80 pixel assignment-mask dilation. Radii are therefore derived automatically rather than selected for individual images. The wider assignment is a fixed robustness margin for boundary uncertainty and token-grid discretization, not a requirement of PromptPack; a user may provide $r_i$ directly. Assignment pixels that extend into $\mathcal K$ may receive regional semantic conditioning, but spatial gating and source restoration still prevent final changes there.

\subsection{Identifying regional dependencies}
\label{supp:regional-dependencies}
Changing a source object can require a corresponding change to its reflection or shadow. For every primary query, we also issue ``reflection of the $q_i$'' and ``shadow of the $q_i$.'' A candidate is accepted if its score is at least $0.50$, overlap with the primary mask is at most $0.10$, area is between $0.05$ and $3$ times the primary area, and minimum distance is at most $0.20$ of the shorter image side. Candidates with mutual IoU above $0.50$ are deduplicated. An accepted dependency is unioned into Modify before constructing the Permit collar and assignment mask. Figure~\ref{fig:sam3_adaptive_showcase} illustrates this process by identifying a swan's reflection as a dependent region that must change consistently with the requested edit.

\subsection{Automatic-grounding evaluation}
\label{supp:automatic-grounding-evaluation}
We screen 11 source-object attribute or material edits drawn from PIE-Bench++ source images. After mask screening, every grounded case receives one PromptPack-Flow FLUX run with seed 0 and a fixed diagnostic setting ($28$ steps, $\tau=4$, and $b=1.3$). We do not tune prompts, seeds, masks, or inference parameters per case. This small screen is a qualitative stress test rather than a benchmark or a replacement for the user-specified-mask evaluation.

\begin{figure*}[t]
  \centering
  \includegraphics[width=0.98\textwidth]{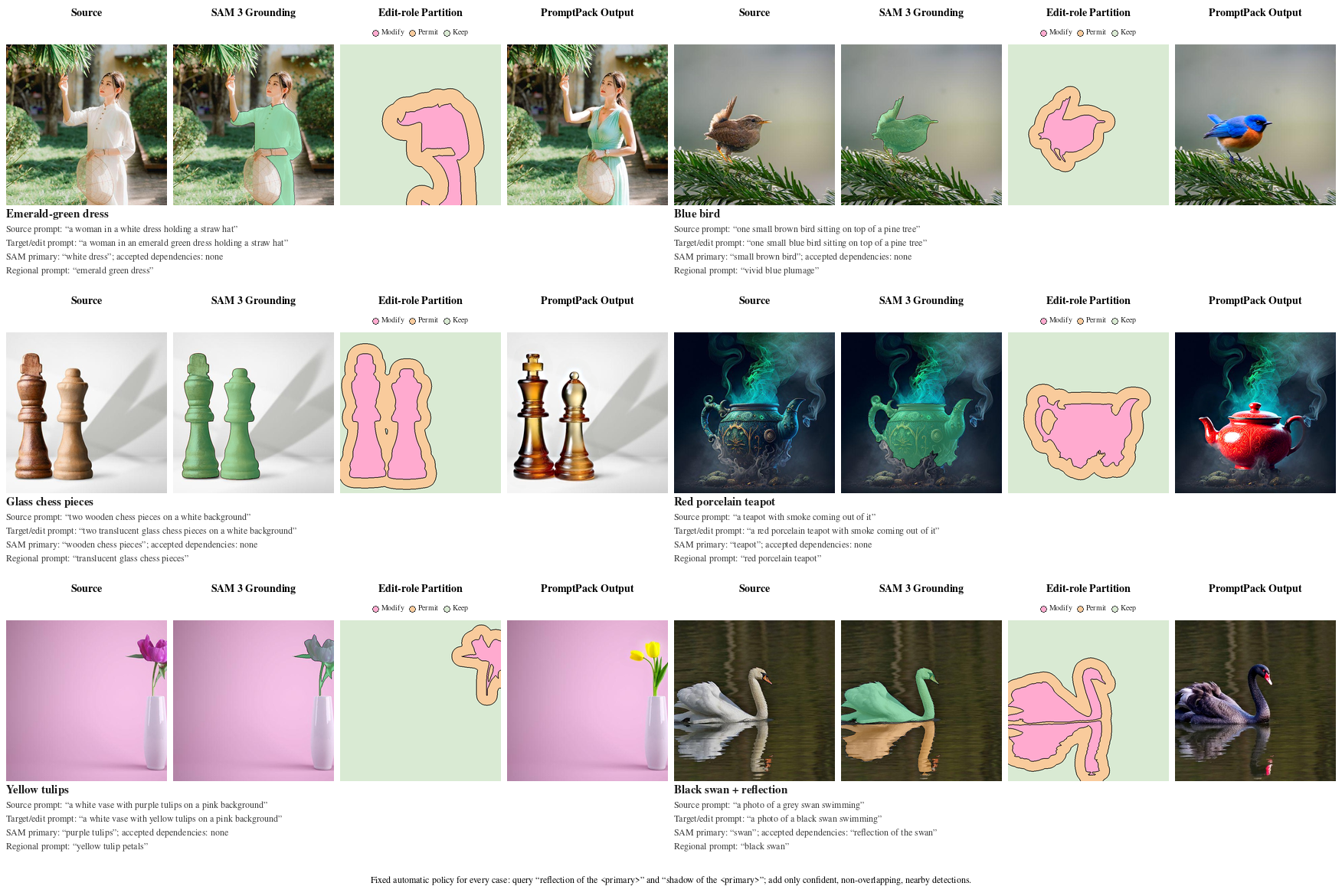}
  \caption{\textbf{Selected automatic-grounding examples.} Each row shows the source, SAM~3 mask, derived edit-role partition, assignment mask, and fixed-seed PromptPack-Flow output. Every example uses the same Edit Guide construction rule. The final row also applies the fixed dependency-query policy, which adds the accepted swan reflection to Modify.}
  \label{fig:sam3_adaptive_showcase}
\end{figure*}

SAM~3 returns an above-threshold mask for 10 of the 11 queries. The screen includes a plural query whose two disconnected chess-piece masks are unioned into one Modify set. Figure~\ref{fig:sam3_adaptive_showcase} shows selected one-shot generations, including successful changes and an accepted dependency mask for the swan reflection. Other grounded cases under-edit the requested attribute despite accurate masks. These outcomes separate two failure sources: automatic segmentation can fail, and a correct spatial control signal cannot make the underlying editor realize a weak semantic change.

\begin{table}[t]
\centering
\small
\setlength{\tabcolsep}{3.5pt}
\begin{tabular}{lccc}
\toprule
\textbf{SAM query} & \textbf{Score} & \textbf{Components} & \textbf{Grounded} \\
\midrule
red apple           & .965          & 1 & Yes \\
white dress         & .958          & 2 & Yes \\
small brown bird    & .980          & 1 & Yes \\
wooden chess pieces & .965, .969    & 2 & Yes \\
dry tree            & --            & -- & No \\
teapot              & .589          & 1 & Yes \\
elephant            & .980          & 1 & Yes \\
cat                 & .968          & 1 & Yes \\
purple tulips       & .758, .764    & 1 & Yes \\
swan                & .985          & 1 & Yes \\
leopard             & .961          & 1 & Yes \\
\bottomrule
\end{tabular}
\caption{\textbf{Complete automatic-grounding screen.} Scores are the selected SAM~3 confidence values; multiple values indicate a union over detections. Components counts connected regions after union. Every grounded case receives one fixed-seed generation using the same Edit Guide construction rule.}
\label{tab:sam3_screening}
\end{table}

\section{Evaluation Protocol}
\label{supp:evaluation-protocol}
This section defines the benchmark split, Edit Guides supplied to each method,
automatic metrics, and human-preference analysis used by the main paper.

\subsection{Dataset split and Edit-Guide annotations}
\label{supp:dataset-edit-guide-annotations}
We evaluate 80 PIE-Bench++ images containing 160 requested concepts. A deterministic 16-image calibration set (indices $0,5,\ldots,75$) is used only to select the two FlowEdit operating points; the remaining 64 images form the disjoint held-out evaluation set.

\paragraph{Regional prompts and Edit Guides.}
PIE-Bench++ provides source and target prompts, with brackets around each
modification concept in the target prompt. For example, source prompt ``a dog
in a space suit'' and target prompt ``a dog [with a scarf] in a space suit
[with flowers in mouth]'' identify two requested modifications. After removing
grammatical filler, these bracketed phrases become the regional prompts
``scarf'' and ``flowers in mouth.'' We manually
annotate the desired location of every requested concept. Each annotation
becomes its regional-prompt assignment; their union initializes Modify, a
dilation collar defines Permit, and all remaining pixels define Keep.

\subsection{Metric definitions and statistical inference}
\label{supp:metric-definitions}
SSIM and LPIPS measure similarity between the edited output and the source
image, CLIP measures global alignment with the target prompt,
$\mathrm{MSE}_{K}$ verifies preservation of the specified Keep set, and ESR,
L-ESR, and WRR measure requested-concept presence, correct regional placement,
and cross-region leakage. Unless a
diagnostic subsection states otherwise, paired 95\% confidence intervals use
100,000 image-clustered bootstrap resamples that keep all concepts from one
image together.

\paragraph{Keep-set MSE.}
To verify protected-pixel preservation, we compute $\mathrm{MSE}_{K}$ over the complete specified Keep set $K=\Omega\setminus(M\cup P)$. PromptPack copies source pixels outside its output support before a Gaussian-smoothed boundary blend (Alg.~\ref{alg:compression_correction}).

\paragraph{ESR Evaluation Protocol.}
Many PIE-Bench++ instructions add target concepts to the source image. For each bracketed concept, we run YOLO-World v2 with confidence threshold 0.1. Detection queries are formed by removing grammatical filler from the bracketed phrases; for example, ``a dog [with a scarf] [and flowers]'' yields ``scarf'' and ``flowers.'' ESR counts a concept when it is detected anywhere. L-ESR requires at least one detection-box center inside that concept's original, undilated assignment mask. WRR counts the concept when any detection center lies inside another concept's assignment but outside its own; a center in an overlap is valid for both and is not counted as wrong. We retain every detection box and center for auditability. These remain open-vocabulary detector measures and do not establish fine-grained attribute correctness.
\subsection{User-study protocol}
\label{supp:user-study}
The user study tests whether people prefer PromptPack to the corresponding base
editor when both receive the same source image and editing request. Candidate
order is randomized, and the instructions ask participants to consider edit
localization, semantic consistency, and overall output quality.

\begin{figure*}[h]
    \centering
    \begin{subfigure}{0.47\linewidth}
    \centering
        \includegraphics[width=\linewidth]{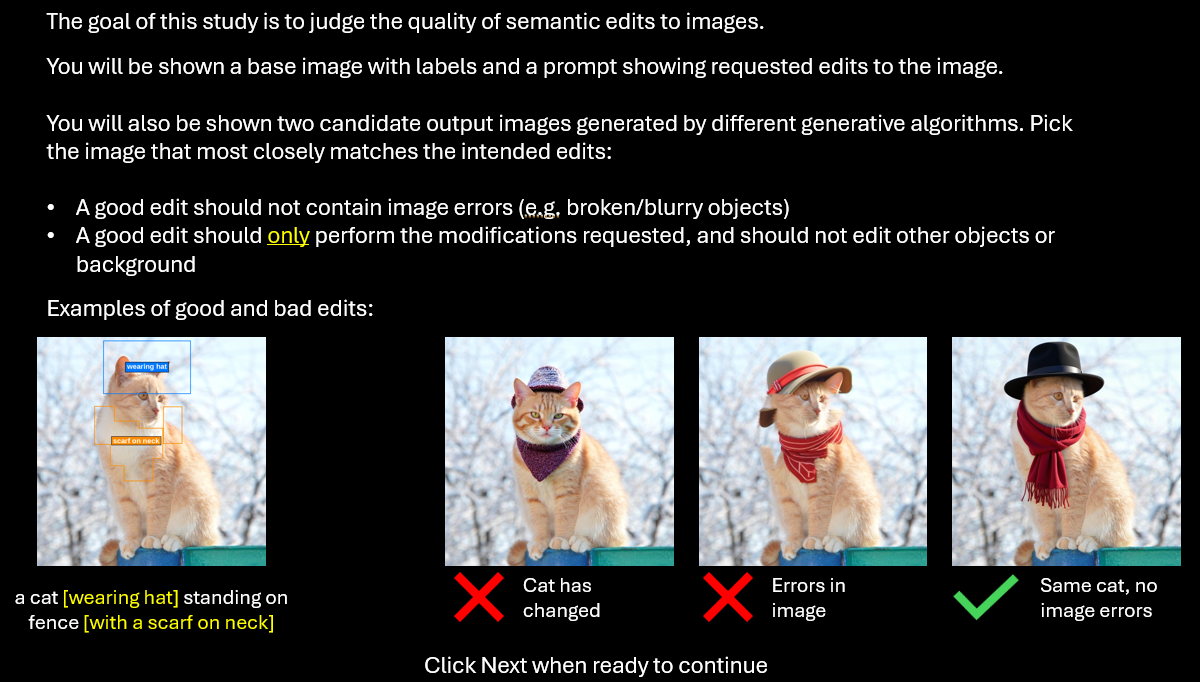}
        \caption{\textbf{Instruction Page} presented to users at the beginning of the study.}
        \centering
        \label{fig:instruction}
    \end{subfigure}%
    \hfill
    \begin{subfigure}{0.47\linewidth}
    \centering
        \includegraphics[width=\linewidth]{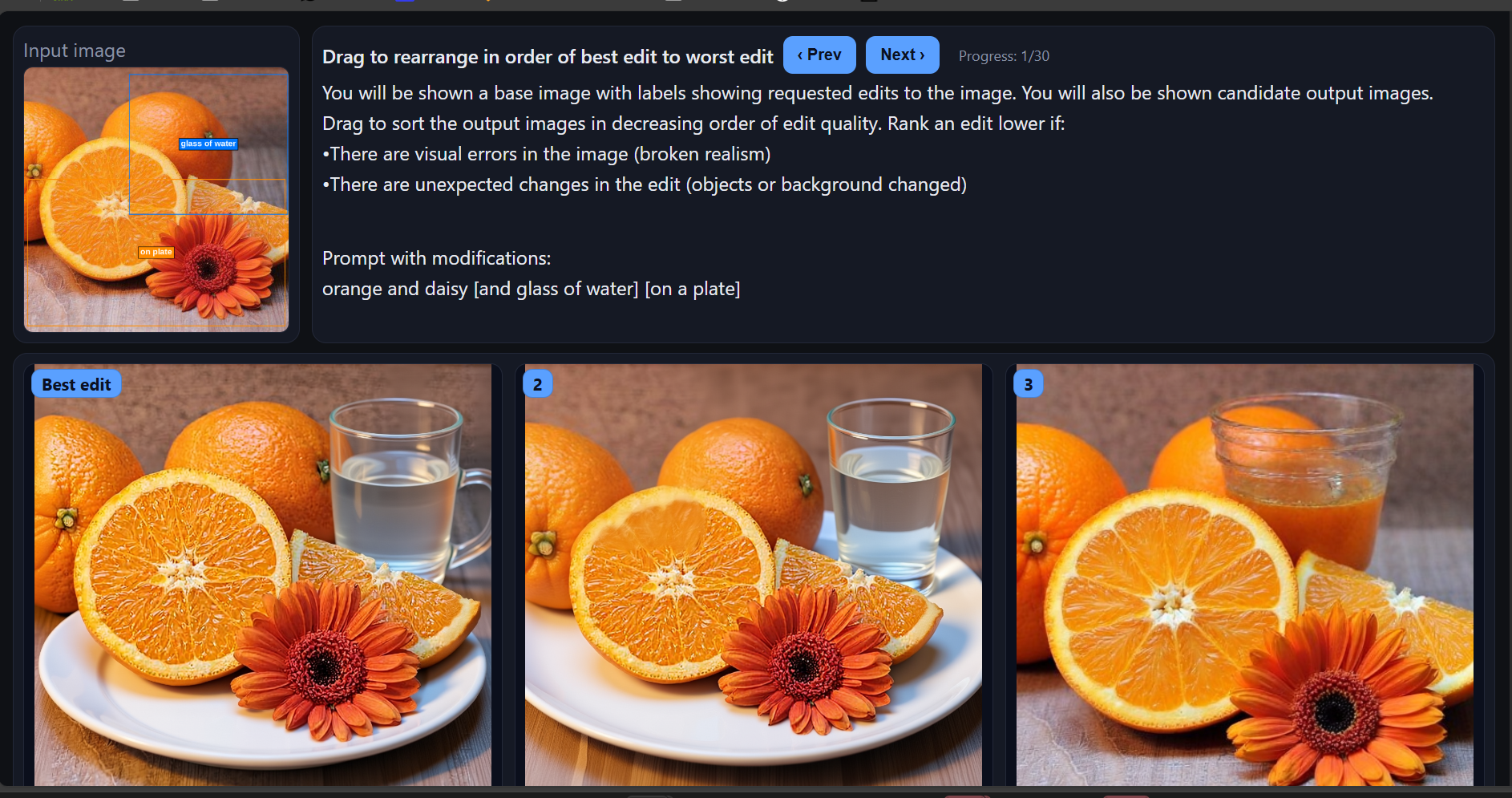}
        \caption{\textbf{Trial screen} for a sampled edit pair present in the user study. Participants are asked to select a winner for each presented pair.}
        \label{fig:trial}
    \end{subfigure}%

    \caption{Screenshots showing details of our user study protocol. On our instruction screen, shown to the users after gathering their informed consent, (a) we ask the users to rate the quality of edits along the axes of edit localization, semantic consistency, and overall output quality. During each trial (b), we show the user an input image, corresponding edit masks and prompts, and two randomly-ordered candidate outputs. The candidate images are generated using one of the three baseline methods (FlowEdit FLUX, FlowEdit SD3, or FLUX Kontext), and its corresponding PromptPack variant. }
    \label{fig:userstudyscreenshots}
\end{figure*}

Our IRB-approved protocol is summarized in the main paper. Figure~\ref{fig:userstudyscreenshots} shows the instructions and trial interface. Participants completed different numbers of trials, so we first compute a preference rate within each participant and pipeline pair, then give each contributing participant equal weight. All participants with at least one recorded comparison are included: 26 for FlowEdit FLUX and FLUX Kontext and 25 for FlowEdit SD3. The analysis contains 769, 768, and 767 judgments, respectively. Confidence intervals use 100,000 participant-clustered bootstrap resamples. We test participant-level rates against 50\% with two-sided Wilcoxon signed-rank tests and apply Holm correction over the three pipelines.

The study was conducted before the final quantitative operating points were
selected and used the fixed configurations in
Table~\ref{tab:user-study-config}. These outputs precede compression correction;
the held-out quantitative evaluation instead uses the settings in
Table~\ref{tab:promptpack_hparams}.

\begin{table}[h!]
\centering
\scriptsize
\setlength{\tabcolsep}{1.8pt}
\caption{PromptPack configurations used to generate the user-study outputs. A dash denotes a FlowEdit-only parameter.}
\label{tab:user-study-config}
\begin{tabular}{lccc}
\toprule
\textbf{Setting} & \shortstack{\textbf{FlowEdit}\\\textbf{FLUX}} & \shortstack{\textbf{FlowEdit}\\\textbf{SD3}} & \shortstack{\textbf{FLUX}\\\textbf{Kontext}} \\
\midrule
Steps & 28 & 50 & 28 \\
Seed & 0 & 0 & 0 \\
Source/target guidance & 1.5/5.5 & 1.5/5.5 & -- \\
Flow range & 0--24 & 0--24 & -- \\
Threshold $\tau$ & 4 & 2 & -- \\
Boost $b$ & 1.3 & 1.2 & -- \\
Dilation begins after & 15 & 15 & 15 \\
Compression correction & No & No & No \\
\bottomrule
\end{tabular}
\end{table}

\begin{figure}[h]
\centering
\includegraphics[width=\linewidth]{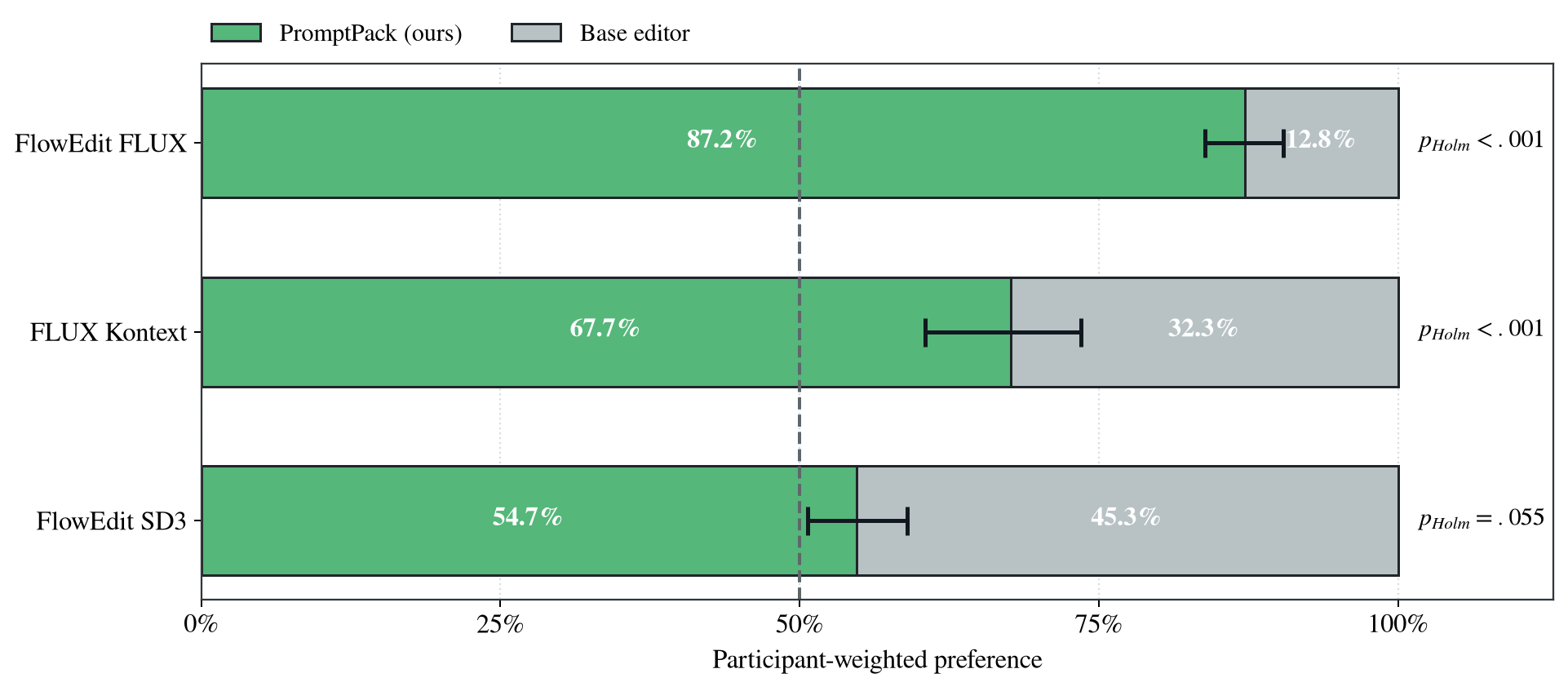}
\caption{Participant-level preference rates between PromptPack and each baseline (F.E. = FlowEdit). Error bars are participant-clustered bootstrap 95\% confidence intervals. The number of contributing participants is 26 for FlowEdit FLUX and FLUX Kontext and 25 for FlowEdit SD3.}
\label{fig:user_study_preference}
\end{figure}

\begin{table}[h!]
\centering
\footnotesize
\setlength{\tabcolsep}{4pt}
\begin{tabular}{l c c c c}
\toprule
\textbf{Model} & \textbf{PP pref. [95\% CI]} & \textbf{Users} & \textbf{Judg.} & \textbf{$p_{\mathrm{Holm}}$} \\
\midrule
FlowEdit FLUX & 87.2 [83.9, 90.4] & 26 & 769 & $<$0.001 \\
FLUX Kontext & 67.7 [60.5, 73.5] & 26 & 768 & $<$0.001 \\
FlowEdit SD3 & 54.7 [50.6, 59.0] & 25 & 767 & 0.055 \\
\bottomrule
\end{tabular}
\caption{Mean participant-level PromptPack preference (percent) with participant-clustered bootstrap 95\% confidence intervals. The SD3 preference is a positive trend but is not significant under the participant-level nonparametric test.}
\label{tab:user_study_detailed}
\end{table}

\begin{figure*}[p]
  \centering
  \includegraphics[width=\textwidth]{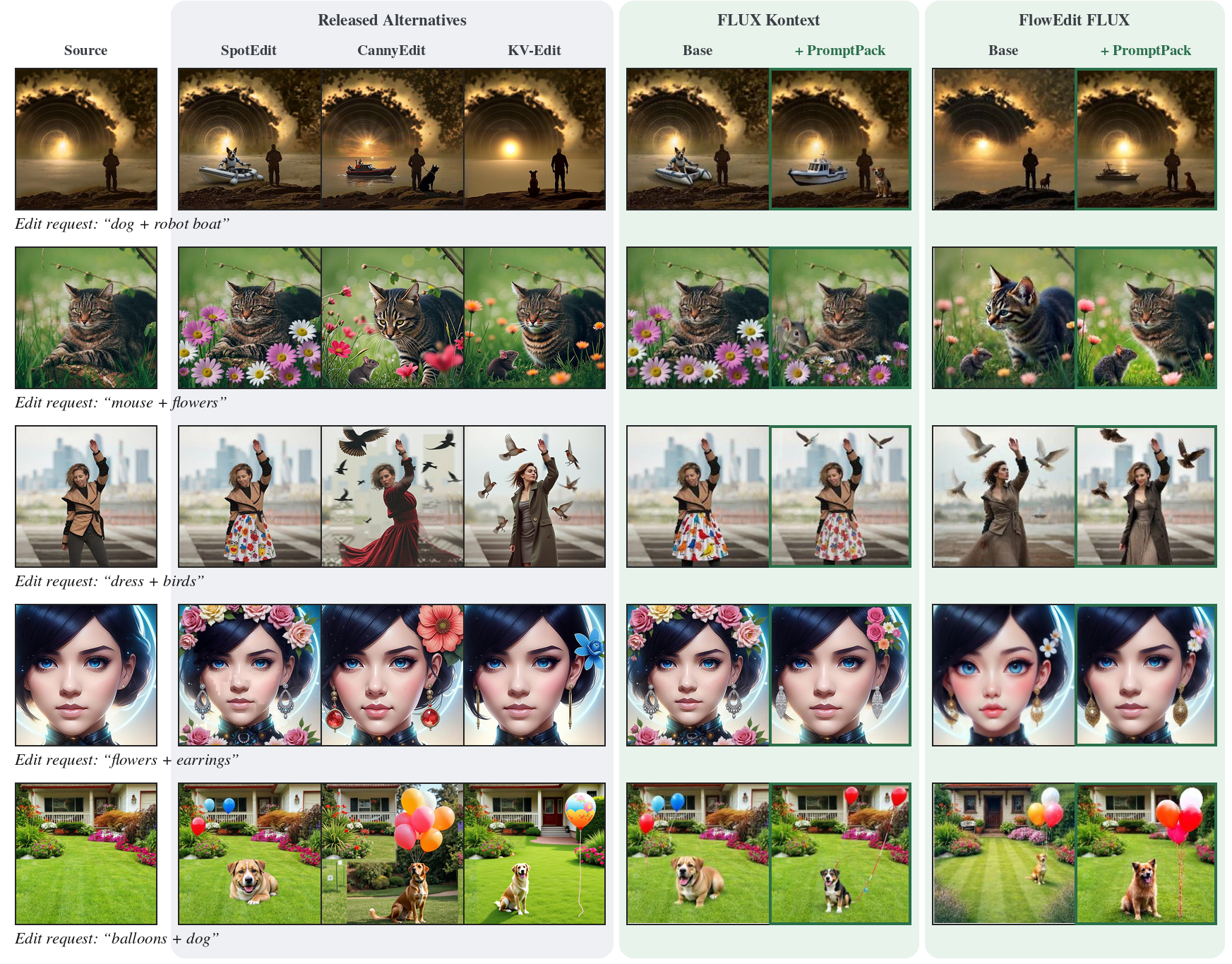}
  \caption{\textbf{Fixed-example comparisons with released alternatives and
  matched base editors.} Each row applies the same held-out PIE-Bench++ edit
  request to every method. The left block provides contextual comparisons with
  SpotEdit, CannyEdit, and KV-Edit; the right blocks show the architecture-matched
  FLUX Kontext and FlowEdit FLUX pairs with and without PromptPack. PromptPack is
  not applied to the released alternatives. Across these examples, PromptPack
  recovers requested concepts omitted by another method or avoids broader
  source changes while realizing the requested edit.}
  \label{fig:released-alternative-qualitative}
\end{figure*}

\begin{figure*}[p]
  \centering
  \includegraphics[width=\textwidth]{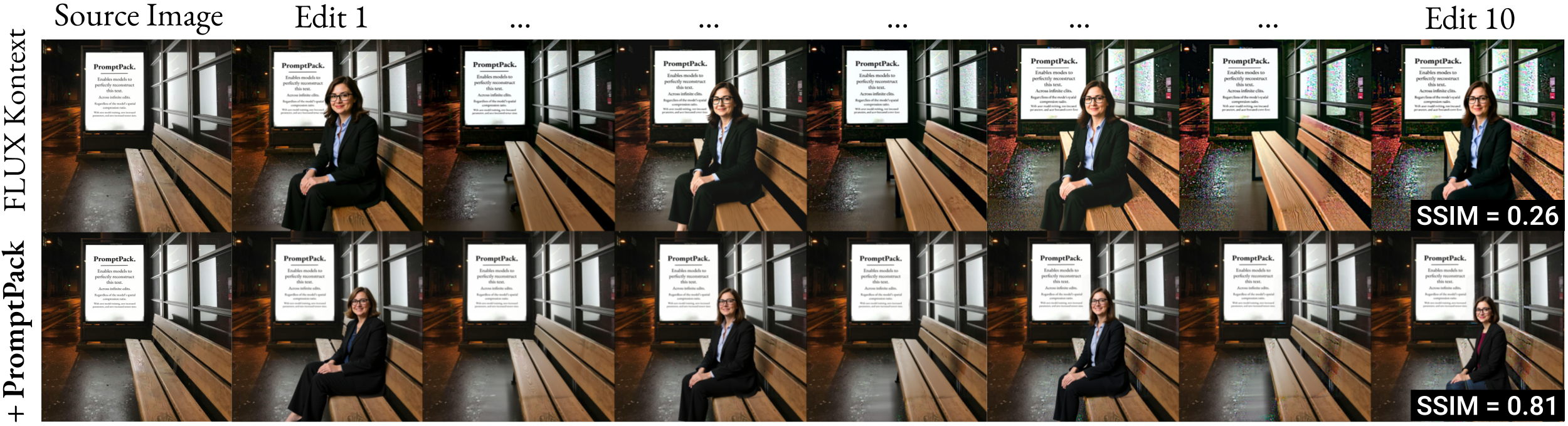}
  \caption{PromptPack (bottom row) reduces edit drift and artifact accumulation over repeated edits compared with FLUX Kontext. Compression correction restores source pixels outside a narrow smoothed transition around the final output support after each turn.}
  \label{fig:multiturn_full}
\end{figure*}

\begin{figure*}[p]
  \centering
  \includegraphics[width=\textwidth]{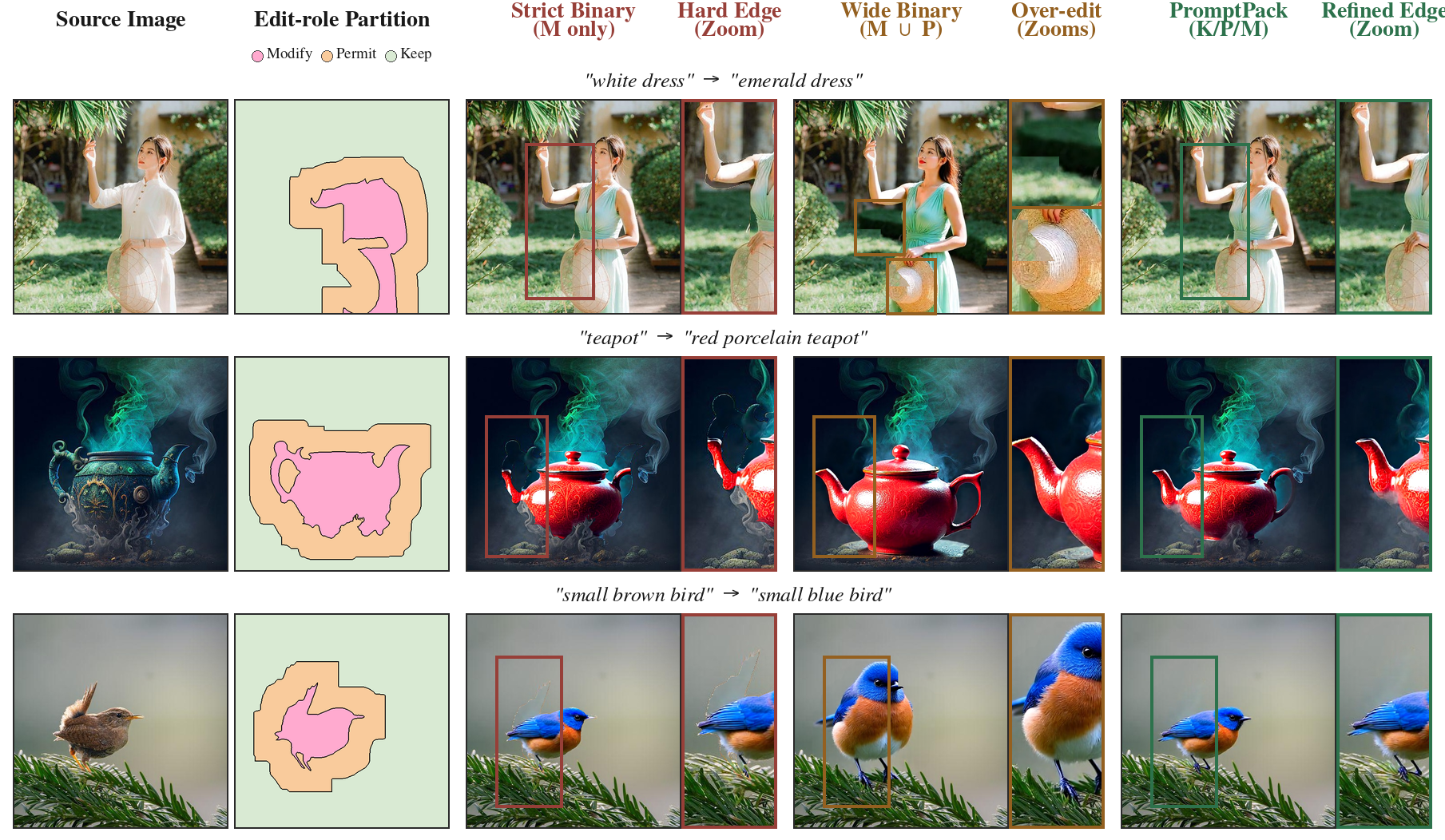}
  \caption{\textbf{Permit removes hard mask-boundary artifacts without the over-editing of a wider binary gate.} In each row, \emph{Modify} is the SAM~3-grounded object mask, \emph{Permit} is a fixed 50-pixel image-space collar, and all remaining pixels are \emph{Keep}. Selected boundary or over-edit regions are enlarged immediately to the right of each output; the two stacked dress crops expose separate changes to the grass and hat. A strict binary gate confines every update to \emph{Modify}, leaving source-colored contours or truncated geometry at the dress, teapot, and bird boundaries. A wide binary gate exposes $M\cup P$ from the first step, removing the hard boundary but changing the woman's appearance and pose, substantially reconstructing the teapot, and changing the bird's scale and pose. PromptPack instead begins from \emph{Modify} and progressively admits \emph{Permit}, allowing each boundary to re-form while retaining the intended object scale, placement, and surrounding content. Source and target prompts, regional assignment, seed, packed embeddings, threshold, and flow boost are identical within each row; only the update gate changes.}
  \label{fig:permit-boundary}
\end{figure*}

\clearpage

\section{Additional Quantitative Evaluation and Robustness}
\label{supp:complete-results}
The main paper reports the confirmatory held-out comparison. This section adds
the full-set aggregates, complete paired uncertainty, comparisons with released
alternatives, sensitivity of the spatial metric to detector choices, and a
capability-level scope comparison. Together, these analyses test whether the
main conclusions depend on a particular split, comparator, or L-ESR definition.

\subsection{Complete benchmark results}
\label{supp:complete-benchmark-results}
Table~\ref{tab:all80-results} reports aggregate metrics over the full 80-image
set, including the 16 images used to select the two FlowEdit operating points.

\begin{table*}[t]
\centering
\small
\setlength{\tabcolsep}{4.2pt}
\caption{Aggregate PIE-Bench++ evaluation over the full 80-image set (160 requested concepts). Semantic rates are percentages. Architecture-matched pairs share the same base editor; secondary baselines expose different controls and are contextual.}
\label{tab:all80-results}
\begin{tabular*}{\textwidth}{@{\extracolsep{\fill}}l c c c c c c c}
\toprule
\textbf{Method} & \textbf{SSIM}$\uparrow$ & \textbf{LPIPS}$\downarrow$ & $\mathbf{MSE}_{K}(10^{-3})\downarrow$ & \textbf{ESR}$\uparrow$ & \textbf{L-ESR}$\uparrow$ & \textbf{WRR}$\downarrow$ & \textbf{CLIP}$\uparrow$ \\
\midrule
FLUX Kontext & .718 & .269 & 8.276 & 58.1 & 34.4 & 14.4 & .319 \\
+ PromptPack & .878 & .162 & .000 & 58.8 & 48.1 & 14.4 & .327 \\
SpotEdit~\cite{qin2026spotedit} & .744 & .249 & 7.870 & 53.8 & 31.9 & 13.8 & .319 \\
\midrule
FlowEdit FLUX & .745 & .277 & 12.769 & 47.5 & 30.0 & 3.8 & .317 \\
+ PromptPack & .818 & .203 & .002 & 55.6 & 47.5 & 1.2 & .322 \\
\midrule
FlowEdit SD3 & .908 & .058 & 1.564 & 30.0 & 11.9 & 4.4 & .300 \\
+ PromptPack & .907 & .087 & .002 & 35.0 & 18.1 & 6.2 & .306 \\
\midrule
FLUX.1 Fill & .700 & .270 & 1.810 & 49.4 & 38.8 & 9.4 & .315 \\
InstructPix2Pix & .620 & .351 & 34.332 & 41.9 & 18.8 & 12.5 & .316 \\
\bottomrule
\end{tabular*}
\end{table*}
\subsection{Held-out paired uncertainty}
\label{supp:heldout-paired-uncertainty}
Table~\ref{tab:heldout-paired-ci} reports the complete paired uncertainty for
the confirmatory 64-image split. Differences are PromptPack minus its
architecture-matched base editor; intervals resample images as clusters so
that all concepts from an image remain together.

\begin{table*}[t]
\centering
\small
\setlength{\tabcolsep}{5pt}
\caption{PromptPack-minus-base changes on the held-out 64 images with paired, image-clustered bootstrap 95\% confidence intervals. $\Delta\mathrm{MSE}_{K}$ is scaled by $10^{-3}$; semantic-rate changes are percentage points.}
\label{tab:heldout-paired-ci}
\begin{tabular}{lccc}
\toprule
\textbf{Metric} & \textbf{FLUX Kontext} & \textbf{FlowEdit FLUX} & \textbf{FlowEdit SD3} \\
\midrule
SSIM & $+.160\ [+.110,+.212]$ & $+.080\ [+.052,+.110]$ & $+.002\ [-.008,+.013]$ \\
LPIPS & $-.106\ [-.141,-.072]$ & $-.082\ [-.113,-.052]$ & $+.027\ [+.018,+.036]$ \\
$\mathrm{MSE}_{K}$ & $-8.095\ [-16.546,-2.553]$ & $-12.707\ [-15.215,-10.357]$ & $-1.645\ [-2.084,-1.266]$ \\
ESR & $+2.3\ [-3.9,+8.7]$ & $+6.3\ [+.8,+11.9]$ & $+2.3\ [-2.3,+7.0]$ \\
L-ESR & $+18.0\ [+10.1,+26.2]$ & $+16.4\ [+8.9,+24.2]$ & $+4.7\ [+.8,+9.1]$ \\
WRR & $-.8\ [-5.5,+3.9]$ & $-3.1\ [-7.0,0.0]$ & $+2.3\ [-.8,+5.9]$ \\
CLIP & $+.008\ [+.002,+.014]$ & $+.004\ [-.001,+.009]$ & $+.005\ [+.001,+.009]$ \\
\bottomrule
\end{tabular}
\end{table*}

\subsection{Quantitative comparisons with released alternatives}
\label{supp:released-alternatives}
Architecture-matched method pairs (e.g., FlowEdit versus FlowEdit with
PromptPack) remain the cleanest measure of PromptPack's effect,
but they do not show where the resulting operating points sit relative to
current released editors. We evaluate SpotEdit, CannyEdit, and KV-Edit using
their released implementations and documented default settings, without
benchmark-specific tuning. These comparisons contextualize preservation and
localized edit success rather than serving as controlled component comparisons.

\begin{figure}[t]
\centering
\includegraphics[width=\linewidth]{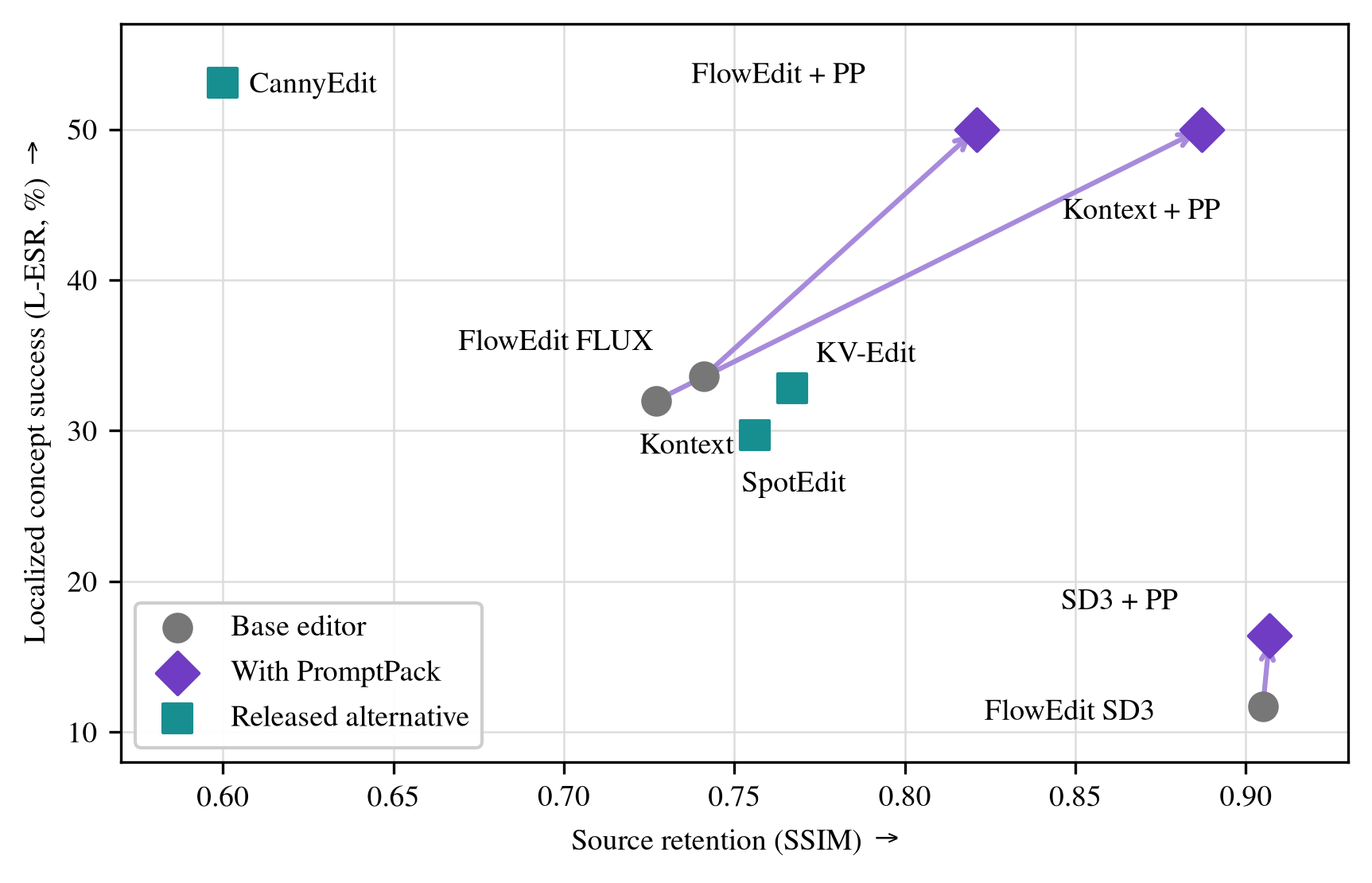}
\caption{\textbf{Preservation--localization context on the held-out split.} PromptPack moves both FLUX base editors toward higher source retention and higher localized concept success. Released alternatives are contextual because they use different controls and editing architectures. CannyEdit has higher global ESR than PromptPack-Flow, but its $+3.1$-point L-ESR difference is unresolved; PromptPack-Flow has significantly higher SSIM and lower wrong-region rate. Arrows connect only architecture-matched PromptPack pairs.}
\label{fig:preservation-localization-pareto}
\end{figure}

Figure~\ref{fig:preservation-localization-pareto} and
Table~\ref{tab:released-alternative-ci} report the held-out comparisons with
SpotEdit, CannyEdit, and KV-Edit; qualitative examples appear in
Fig.~\ref{fig:released-alternative-qualitative}.

\begin{table*}[t]
\centering
\small
\setlength{\tabcolsep}{4.5pt}
\caption{Paired held-out differences against released current alternatives, reported as PromptPack minus the named method with image-clustered bootstrap 95\% confidence intervals. SpotEdit is compared with PromptPack applied to FLUX Kontext; CannyEdit and KV-Edit are compared with PromptPack-Flow. Semantic differences are percentage points. These are contextual, not architecture-matched, contrasts.}
\label{tab:released-alternative-ci}
\begin{tabular}{lccccc}
\toprule
\textbf{Alternative} & $\boldsymbol{\Delta}$\textbf{SSIM} & $\boldsymbol{\Delta}$\textbf{LPIPS} & $\boldsymbol{\Delta}$\textbf{ESR} & $\boldsymbol{\Delta}$\textbf{L-ESR} & $\boldsymbol{\Delta}$\textbf{WRR} \\
\midrule
SpotEdit & $+.131\ [+.084,+.182]$ & $-.086\ [-.119,-.054]$ & $+7.0\ [+1.5,+12.8]$ & $+20.3\ [+13.0,+27.9]$ & $-1.6\ [-6.1,+2.4]$ \\
CannyEdit & $+.220\ [+.198,+.243]$ & $-.214\ [-.238,-.190]$ & $-6.3\ [-12.4,-.8]$ & $-3.1\ [-10.9,+4.7]$ & $-7.8\ [-12.4,-3.8]$ \\
KV-Edit & $+.054\ [+.034,+.075]$ & $-.040\ [-.062,-.019]$ & $+9.4\ [+1.6,+16.9]$ & $+17.2\ [+9.3,+25.2]$ & $-7.8\ [-12.5,-3.8]$ \\
\bottomrule
\end{tabular}
\end{table*}

\subsection{Spatial-metric sensitivity}
\label{supp:spatial-metric-sensitivity}
The Localized Edit Success Ratio (L-ESR) depends on both an open-vocabulary
detector and a rule for deciding
whether a detection belongs to an assignment region. We therefore test whether
the architecture-matched direction changes under stricter detector thresholds,
alternative overlap rules, an independent detector, and a blinded human audit.

We re-score the saved YOLO-World boxes without rerunning either the detector or
the editors. Table~\ref{tab:metric-sensitivity} gives PromptPack-minus-base
L-ESR differences. All directions remain positive. Absolute rates decline as
the confidence threshold increases; the permissive any-overlap rule narrows
the FLUX Kontext difference because very large boxes can touch both assignments.

\begin{table}[t]
\centering
\small
\setlength{\tabcolsep}{3.5pt}
\caption{L-ESR sensitivity in percentage points. Confidence rows use box-center membership; geometry rows use confidence $.1$. ``Box 25/50'' requires that fraction of the detection box to lie in the assignment mask.}
\label{tab:metric-sensitivity}
\begin{tabular}{lrrr}
\toprule
\textbf{Criterion} & \textbf{FLUX Kontext} & \textbf{FLUX} & \textbf{SD3} \\
\midrule
Confidence $.10$ & +13.8 & +17.5 & +6.3 \\
Confidence $.20$ & +10.6 & +18.1 & +4.4 \\
Confidence $.30$ & +12.5 & +16.3 & +5.0 \\
Confidence $.50$ & +15.0 & +12.5 & +3.8 \\
\midrule
Any overlap & +2.5 & +13.1 & +5.6 \\
Box 25\% & +10.6 & +15.6 & +7.5 \\
Box 50\% & +16.3 & +18.1 & +6.3 \\
\bottomrule
\end{tabular}
\end{table}

\paragraph{Independent-detector cross-check.}
\phantomsection\label{supp:independent-detector}
Grounding DINO-Tiny~\cite{liu2024groundingdino} independently processes all 64
held-out images (128 requested concepts), one inference per requested phrase,
and we apply the same box-center membership rule to the
same original assignment masks. Table~\ref{tab:grounding-dino-sensitivity}
reports L-ESR across detector thresholds. PromptPack-Flow exceeds FlowEdit by
$8.6$--$14.1$ points, and all four paired image-clustered intervals exclude
zero. Grounding DINO's global ESR saturates at 96.9--99.2\%, illustrating why
absolute rates should not be compared across detectors. PromptPack-minus-CannyEdit
L-ESR ranges from $-2.3$ to $+3.1$ points, with every interval including zero.
This cross-check establishes detector-family robustness for regional placement,
not fine-grained attribute verification.

\begin{table}[t]
\centering
\scriptsize
\setlength{\tabcolsep}{2pt}
\caption{Grounding DINO-Tiny L-ESR on all 64 held-out images. The final column is PromptPack-Flow minus FlowEdit FLUX with paired image-clustered 95\% confidence intervals; rates and differences are percentage points.}
\label{tab:grounding-dino-sensitivity}
\begin{tabular}{lrrrr}
\toprule
\textbf{Conf.} & \textbf{Flow} & \textbf{PP} & \textbf{Canny} & $\boldsymbol{\Delta}$\textbf{ PP--Flow} \\
\midrule
.15 & 79.7 & 88.3 & 90.6 & $+8.6\ [+1.6,+15.9]$ \\
.20 & 77.3 & 85.9 & 84.4 & $+8.6\ [+.8,+16.3]$ \\
.25 & 71.9 & 82.8 & 82.0 & $+10.9\ [+3.1,+19.2]$ \\
.30 & 68.0 & 82.0 & 78.9 & $+14.1\ [+5.5,+22.7]$ \\
\bottomrule
\end{tabular}
\end{table}

\paragraph{Method-blinded human audit.}
\phantomsection\label{supp:human-audit}
To validate the automatic L-ESR labels, we ask human annotators whether each
requested concept appears inside its assigned region in outputs from FlowEdit
FLUX, FlowEdit FLUX with PromptPack, and CannyEdit. We audit 16 held-out images
selected by a fixed simple-random draw (seed 20260824). One of 31 candidate
concepts has an empty assignment and is excluded before scoring, leaving 30
concepts and 90 output judgments. For each concept, the three methods are independently permuted as
A/B/C. Three annotators (including one author), blinded to method identity, see
each output with only the relevant assignment boundary and label the named
concept as mostly inside, mostly outside, absent, or uncertain; roughly
half-in/half-out cases are uncertain. We retain uncertainty as its own category
and form consensus only when at least two annotators give the same four-way
label. Of 90 outputs, 71 (78.9\%) are unanimous and 86 (95.6\%) have a
four-way majority. Fleiss' $\kappa=.674$ and nominal Krippendorff
$\alpha=.675$ across four labels; for the metric's central inside-versus-not-
inside distinction, treating uncertainty as missing gives $\alpha=.886$.

Each annotator independently finds more inside placements for PromptPack-Flow
(24, 26, and 23) than FlowEdit (16, 15, and 12), and no annotator assigns any
PromptPack output outside its requested region. Their paired definite screens
give 7, 10, and 7 PromptPack wins, respectively, with no FlowEdit wins (exact
$p=.016$, $.002$, and $.016$). Majority consensus (Table~\ref{tab:manual-lesr-audit})
gives 25 inside and zero outside for PromptPack, versus 14 and two for FlowEdit.
On the 25 concepts definite for both methods, PromptPack wins seven, loses none,
and improves inside rate by 28.0 points (source-image-clustered bootstrap 95\%
CI $[+12.5,+44.0]$; two-sided exact McNemar $p=.016$). PromptPack and CannyEdit
remain unresolved: over 26 jointly definite concepts, PromptPack wins three and
loses four (difference $-3.8$ points, CI $[-29.6,+19.2]$, $p=1.0$).

On the same subset, YOLO-World reports L-ESR of 33.3\%, 60.0\%, and 63.3\% for
FlowEdit, PromptPack, and CannyEdit. Against the 81 definite majority-consensus
labels, automated L-ESR has 40 true positives, two false positives, 23 false
negatives, and 16 true negatives: 95.2\% precision and 63.5\% recall. Thus it is
a conservative indicator whose architecture-matched direction agrees with the
human audit, rather than exhaustive semantic judgment.

\begin{table}[t]
\centering
\scriptsize
\setlength{\tabcolsep}{2.6pt}
\caption{Three-annotator, method-blinded spatial audit on 30 fixed random held-out concepts. Human entries are majority-consensus counts among 30 outputs per method; ``Unres.'' denotes three distinct labels. Automatic L-ESR uses the same outputs.}
\label{tab:manual-lesr-audit}
\begin{tabular}{lrrrrrr}
\toprule
\textbf{Method} & \textbf{Inside} & \textbf{Outside} & \textbf{Absent} & \textbf{Uncert.} & \textbf{Unres.} & \textbf{Auto L-ESR} \\
\midrule
FlowEdit FLUX & 14 & 2 & 9 & 2 & 3 & 33.3 \\
$+$ PromptPack & \textbf{25} & \textbf{0} & 4 & 1 & 0 & 60.0 \\
CannyEdit & 24 & 1 & 2 & 2 & 1 & 63.3 \\
\bottomrule
\end{tabular}
\end{table}

\paragraph{Regional Prompt Retrieval.}
\phantomsection\label{supp:regional-prompt-retrieval}
To test whether each edited region contains its assigned concept without
relying on an object detector, we construct a CLIP ViT-B/32~\cite{radford2021learning}
cosine-similarity matrix between output crops defined by the original,
undilated assignment masks and all regional phrases within each multi-region
image. For each region, crop-to-prompt retrieval asks whether
its assigned phrase has maximum similarity; prompt-to-region retrieval asks
the converse. We report their mean top-1 accuracy and the mean correct-pair
margin over the strongest competing phrase or crop. The ordinary view is the
smallest square enclosing the assignment; the masked view replaces pixels
outside the assignment with neutral gray. This within-image retrieval avoids
cross-image prompt-difficulty confounds and follows the use of regional CLIP
similarity in RegionRoute~\cite{chen2026regionroute}.

The fixed screen contains all 60 held-out images with at least two valid
assignments (123 regions); one empty annotation and three single-region images
cannot define retrieval. Table~\ref{tab:regional-prompt-retrieval} shows that
both FLUX integrations increase the continuous correct-pair margin under both
views. For FLUX Kontext, the paired box and masked margin changes are $+.0054$
($[+.0005,+.0107]$) and $+.0079$ ($[+.0034,+.0127]$); for FlowEdit FLUX they
are $+.0045$ ($[+.0006,+.0085]$) and $+.0068$
($[+.0025,+.0114]$). SD3 changes are unresolved. Discrete top-1 differences
are mostly unresolved, so we use this as corroborating regional association,
not a replacement for L-ESR or human semantic judgment.

\begin{table}[t]
\centering
\scriptsize
\setlength{\tabcolsep}{2.3pt}
\caption{Bidirectional Regional Prompt Retrieval (RPR) on the complete valid multi-region held-out set. Accuracy is a percentage; margin is CLIP cosine similarity.}
\label{tab:regional-prompt-retrieval}
\begin{tabular}{lrrrr}
\toprule
& \multicolumn{2}{c}{\textbf{Box crop}} & \multicolumn{2}{c}{\textbf{Masked crop}} \\
\cmidrule(lr){2-3}\cmidrule(lr){4-5}
\textbf{Method} & \textbf{RPR} & \textbf{Margin} & \textbf{RPR} & \textbf{Margin} \\
\midrule
FLUX Kontext & 63.0 & .0124 & 67.5 & .0186 \\
$+$ PromptPack & 66.3 & .0178 & 72.0 & .0265 \\
FlowEdit FLUX & 65.9 & .0148 & 75.2 & .0267 \\
$+$ PromptPack & 69.1 & .0193 & 77.6 & .0335 \\
FlowEdit SD3 & 58.1 & .0095 & 62.2 & .0109 \\
$+$ PromptPack & 56.9 & .0083 & 64.6 & .0103 \\
\bottomrule
\end{tabular}
\end{table}

\subsection{Capability comparison and scope}
\label{supp:capability-scope}
Quantitative comparisons cannot cleanly isolate method capabilities when
editors require different inputs or target different tasks. We therefore use a
separate literature-based table to state which aspects of PromptPack's interface
are documented by each closely related method; this table is descriptive, not
an empirical ranking.

Table~\ref{tab:capability-comparison} summarizes only capabilities established
by each cited paper. ``NR'' means not reported or not established in that work;
it does not assert that an implementation is incapable of the feature.

\begin{center}
\begin{minipage}{\columnwidth}
\centering
\scriptsize
\setlength{\tabcolsep}{1.3pt}
\renewcommand{\arraystretch}{0.96}
\captionof{table}{Documented capabilities of closely related spatial editing and regional-conditioning methods. ``Multi'' denotes multiple regional prompts, and ``Indep.'' denotes independently assigned regions. Backbone and task appear below each method.}
\label{tab:capability-comparison}
\begin{tabularx}{\columnwidth}{@{}>{\raggedright\arraybackslash}X c c c c c@{}}
\toprule
\textbf{Method / task} & \textbf{Train} & \textbf{Multi} & \textbf{Indep.} & \textbf{K/P/M} & \textbf{Restore} \\
\midrule
\shortstack[l]{TFRP~\cite{tfrp2024}\\[-1pt]\tiny FLUX text-to-image} & No & Yes & Yes & NR & NR \\
\shortstack[l]{ObjectAdd~\cite{zhang2024objectadd}\\[-1pt]\tiny SD-v1.4 object insertion} & No & NR & NR & NR & NR \\
\shortstack[l]{RegionRoute~\cite{chen2026regionroute}\\[-1pt]\tiny FLUX Kontext style transfer} & Yes & NR & NR & NR & NR \\
\shortstack[l]{CARE-Edit~\cite{wang2026careedit}\\[-1pt]\tiny FLUX contextual editing} & Yes & NR & NR & NR & NR \\
\shortstack[l]{BRIDGE~\cite{xiong2026bridge}\\[-1pt]\tiny Coarse-mask local editing} & Yes & NR & NR & NR & NR \\
\shortstack[l]{SpotEdit~\cite{qin2026spotedit}\\[-1pt]\tiny FLUX Kontext selective editing} & No & NR & NR & NR & NR \\
\shortstack[l]{KV-Edit~\cite{zhu2025kvedit}\\[-1pt]\tiny FLUX mask-guided editing} & No & NR & NR & NR & NR \\
\shortstack[l]{CannyEdit~\cite{xie2025cannyedit}\\[-1pt]\tiny FLUX local--global editing} & No & Yes & Yes & NR & NR \\
\shortstack[l]{\textbf{PromptPack}\\[-1pt]\tiny FLUX/SD3 FlowEdit; FLUX Kontext} & \textbf{No} & \textbf{Yes} & \textbf{Yes} & \textbf{Yes} & \textbf{Yes} \\
\bottomrule
\end{tabularx}
\end{minipage}
\end{center}

\long\def\PromptPackDesignAblations{%
\section{Design Choice Ablations}
\label{supp:design-ablations}
This section asks which design choices produce regional control and which trade
edit strength against preservation. Controlled screens vary conditioning,
capacity, gating, and FlowEdit updates while holding all other inputs fixed.

\subsection{Fixed-length prompt packing and regional routing}
\label{supp:packing-routing}
To isolate fixed-length packing from the other PromptPack components, we fix
all 64 held-out images (128 requested concepts), seed 0, FlowEdit FLUX, the
final $(\tau,b)=(4,1.6)$ operating point, the progressive
K/P/M gate, assignment masks, and source correction. The ordinary condition
uses the original 512-position global target sequence without a regional
conditioning path. A packed-unrouted control inserts the independently encoded
regional content into available positions but restores ordinary unmasked
attention. The expanded condition concatenates one independently
encoded full 512-position sequence for the global target and for every regional
prompt, then applies the same regional routing as PromptPack. This is a
TFRP-style conditioning-topology diagnostic adapted inside the editor, not a
reproduction of TFRP's text-to-image generation method. The packed condition
instead removes regional padding and inserts the same regional content
embeddings into the global sequence's available positions, preserving length
512. Three cases contain one regional prompt (expanded length 1024), 58 contain
two (1536), and three contain three (2048), giving mean length 1536. The screen
is the complete confirmatory split rather than a selected subset.

\begin{table*}[t]
\centering
\small
\setlength{\tabcolsep}{4.5pt}
\caption{Complete held-out conditioning-topology screen. Rates are percentages; $L$ is mean sequence length. Timing and peak CUDA allocation are means on an RTX PRO 6000.}
\label{tab:packing-topology-full}
\begin{tabular}{lccccccc}
\toprule
\textbf{Conditioning} & $\mathbf{L}$ & \textbf{SSIM}$\uparrow$ & \textbf{LPIPS}$\downarrow$ & \textbf{ESR}$\uparrow$ & \textbf{L-ESR}$\uparrow$ & \textbf{WRR}$\downarrow$ & \textbf{s / GiB}$\downarrow$ \\
\midrule
Ordinary joint & 512 & .836 & .180 & 58.6 & 45.3 & 3.1 & 8.21 / 37.55 \\
Packed, no routing & 512 & .824 & .195 & 57.8 & 43.0 & 7.8 & 8.13 / 37.74 \\
Expanded routed & 1536 & .813 & .206 & 58.6 & 46.9 & 6.2 & 12.96 / 47.99 \\
Packed routed (ours) & 512 & .821 & .198 & 58.6 & \textbf{50.0} & \textbf{1.6} & 9.70 / 37.74 \\
\bottomrule
\end{tabular}
\end{table*}

Paired image-clustered intervals separate the component effects. Packed routing
versus packed unmasked conditioning changes L-ESR by $+7.0$ points
($[+.8,+13.2]$), WRR by $-6.3$ ($[-10.6,-2.4]$), and SSIM by $-.004$
($[-.011,+.002]$). Packed versus expanded routing changes L-ESR by $+3.1$
($[-4.5,+10.7]$), WRR by $-4.7$ ($[-8.6,-1.6]$), and SSIM by $+.007$
($[-.002,+.017]$). Packed routing versus the ordinary joint prompt changes
L-ESR by $+4.7$ ($[-.8,+10.3]$) and SSIM by $-.015$
($[-.023,-.008]$). We therefore attribute the reliable spatial gain to routing;
fixed-length packing preserves that routing while avoiding expanded sequence
cost, rather than uniformly improving every image metric.

\subsection{Prompt-capacity stress test}
\label{supp:capacity-stress}
The benchmark prompts all fit within the native conditioning length
(Sec.~\ref{supp:prompt-capacity}), so they do not reveal how PromptPack behaves
when fewer positions remain available. We simulate constrained capacity by
retaining at most 4 or 8 regional embeddings and compare both settings with the
full regional content on a fixed 16-image FlowEdit FLUX subset.

\begin{table}[t]
\centering
\scriptsize
\setlength{\tabcolsep}{2.0pt}
\caption{Regional-token capacity stress on 16 fixed held-out FlowEdit FLUX cases (31 concepts). ``Retained'' is the mean regional embeddings and mean fraction of full regional content. Masked RPR uses 14 valid multi-region cases.}
\label{tab:capacity-stress}
\begin{tabular}{lrrrrr}
\toprule
\textbf{Budget} & \textbf{Retained} & \textbf{SSIM} & \textbf{L-ESR} & \textbf{RPR} & \textbf{CLIP} \\
\midrule
4 slots & 4.0 (48\%) & .869 & 38.7 & 76.8 & .322 \\
8 slots & 7.6 (87\%) & .860 & 38.7 & 87.5 & .324 \\
Full & 9.1 (100\%) & .858 & 45.2 & 89.3 & .326 \\
\bottomrule
\end{tabular}
\end{table}

Relative to four slots, full regional text changes L-ESR by $+6.5$ points
($[0,+16.1]$) and masked RPR by $+12.5$ points ($[-1.8,+28.6]$), while SSIM
falls by $.011$ ($[-.018,-.005]$) because the full condition performs stronger
edits. Eight slots retain 87\% of regional embeddings on average and remain
close to full: the full-minus-eight differences are $+6.5$ L-ESR points
($[0,+16.1]$), $+1.8$ masked-RPR points ($[-7.1,+10.7]$), and $-.002$ SSIM
($[-.0049,-.0002]$). Thus moderate slot pressure is stable in this small screen,
whereas severe truncation trends toward under-editing. This is a stress
diagnostic for one integration, not a general over-capacity guarantee or a
recommended truncation policy.

\subsection{Three-state spatial gating versus binary masks}
\label{supp:three-state-gating}
An ordinary binary mask must either tightly restrict the edit or expose its entire allowable support from the beginning. Figure~\ref{fig:permit-boundary} isolates the resulting boundary-coherence tradeoff, while Fig.~\ref{fig:kpm-binary} compares the broader choices with PromptPack's progressive three-state gate and with an external inpainting baseline.

\begin{figure*}[t]
  \centering
  \includegraphics[width=\textwidth]{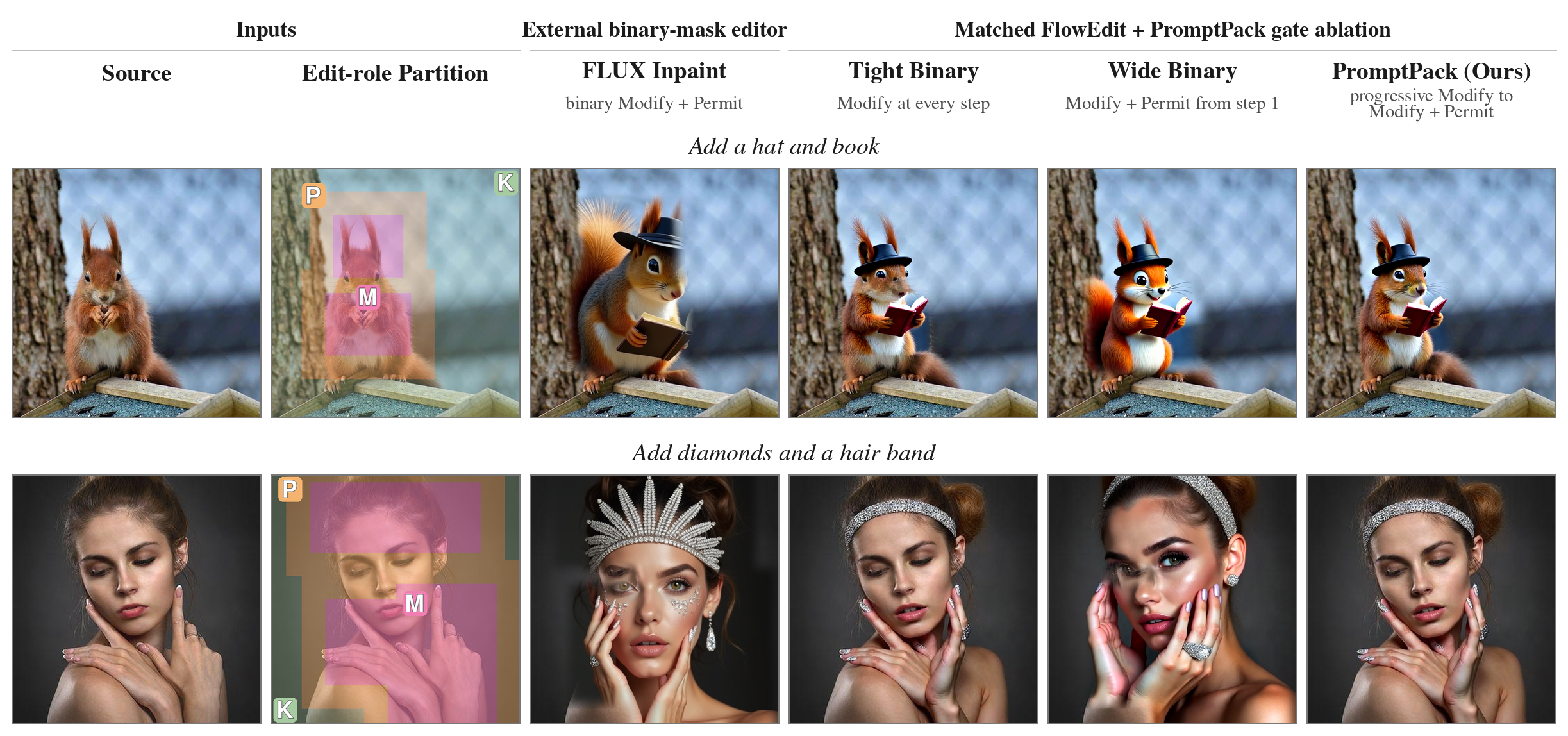}
  \caption{\textbf{Why use a three-state edit-role gate?} FLUX Inpaint receives the full target prompt and the binary editable support $M\cup P$. Within each row, prompts, packed embeddings, regional prompt assignments, seed, flow threshold, and boost are fixed across the three rightmost columns; only the FlowEdit update gate changes. A tight binary gate holds updates to $M$, whereas a wide gate exposes $M\cup P$ from the first editing step. PromptPack instead begins at $M$ and progressively expands through $P$. Inpainting and the wide gate reconstruct substantially more of the subject; the tight gate limits the edit extent; progressive \emph{Keep}/\emph{Permit}/\emph{Modify} control preserves source identity while allowing a fuller local edit. Table~\ref{tab:kpm-quantitative} quantifies this controlled screen.}
  \label{fig:kpm-binary}
\end{figure*}

Before generation, we fixed a mechanical selection rule: the first 20 held-out
indices in numerical order, excluding all calibration images. With seeds 0 and
1, the screen contains 40 case--seed samples and 78 requested-concept trials.
All FlowEdit conditions use the final $(\tau,b)=(4,1.6)$ setting, identical
packed prompts and assignments, and differ only in the spatial update gate.
FLUX Inpaint receives the same source, target prompt, seed, and widest $M\cup P$
support. Figure~\ref{fig:kpm-binary} shows two fixed-sample cases selected for
visual clarity after the quantitative screen was complete.

\begin{table}[t]
\centering
\small
\setlength{\tabcolsep}{3pt}
\caption{Mechanically selected held-out three-state screen. Raw $\mathrm{MSE}_{K}$ is measured before source restoration; SSIM, ESR, and L-ESR use the final output. $\mathrm{MSE}_{K}$ is scaled by $10^{-3}$ and detection rates are percentages.}
\label{tab:kpm-quantitative}
\begin{tabular}{lcccc}
\toprule
\textbf{Condition} & \textbf{SSIM}$\uparrow$ & $\mathbf{MSE}^{raw}_{K}\downarrow$ & \textbf{ESR}$\uparrow$ & \textbf{L-ESR}$\uparrow$ \\
\midrule
Tight ($M$) & .858 & .658 & 56.2 & 40.0 \\
Wide ($M\cup P$) & .773 & 1.135 & 62.5 & 45.0 \\
K/P/M & .849 & .666 & 57.5 & 42.5 \\
FLUX inpaint & .647 & 1.116 & 53.8 & 28.7 \\
\bottomrule
\end{tabular}
\end{table}

Relative to the tight gate, K/P/M gains 2.5 L-ESR points (case-clustered
95\% CI $[0,6.3]$) while changing SSIM by $-.009$ ($[-.012,-.006]$).
Relative to the wide gate, it improves SSIM by $.076$ ($[.056,.097]$) and
reduces raw Keep error by $.469\times10^{-3}$
($[.261,.705]\times10^{-3}$); its $-2.5$ L-ESR-point difference is unresolved
($[-11.3,6.3]$). Against inpainting, K/P/M improves SSIM by $.202$
($[.161,.248]$) and L-ESR by 13.8 points ($[2.5,25.0]$). These results establish
a preservation--editability compromise over this fixed screen, not universal
superiority or the value of user-authored asymmetric Permit regions.

\subsection{Asymmetric Permit regions}
\label{supp:asymmetric-permit}
The main benchmark mechanically derives a symmetric Permit collar from each
placement box. To exercise the more expressive interface, we predeclare 12
held-out single-concept tasks whose requested object has one unambiguous
cardinal attachment direction in the source (e.g., a hat toward a head or a
bird toward an apple). Selection uses only source images and annotations before
generation. Every condition shares FlowEdit FLUX, the packed prompt, assignment,
seed 0, $(\tau,b)=(4,1.6)$, and source correction. The directional guide retains
a one-token seam collar in every direction and extends Permit to three tokens
only toward the declared attachment; the symmetric guide extends three tokens
in every direction. Tight $M$ and wide binary $M\cup P$ controls complete the
screen.

\begin{table}[t]
\centering
\small
\setlength{\tabcolsep}{3.5pt}
\caption{Predeclared 12-task asymmetric-Permit screen. Support is the fraction of image tokens allowed to change; semantic rates are percentages.}
\label{tab:asymmetric-permit}
\begin{tabular}{lrrrr}
\toprule
\textbf{Gate} & \textbf{Support}$\downarrow$ & \textbf{SSIM}$\uparrow$ & \textbf{ESR}$\uparrow$ & \textbf{L-ESR}$\uparrow$ \\
\midrule
Tight $M$ & 15.2 & .942 & 33.3 & 33.3 \\
Wide $M\cup P$ & 29.2 & .903 & 33.3 & 25.0 \\
Symmetric K/P/M & 29.2 & .925 & 33.3 & 33.3 \\
Directional K/P/M & 23.2 & .929 & 33.3 & 33.3 \\
\bottomrule
\end{tabular}
\end{table}

Relative to symmetric K/P/M, the directional guide reduces support by 6.0
percentage points (paired 95\% CI $[-7.2,-4.8]$) and improves SSIM by $+.004$
($[+.002,+.006]$); ESR and L-ESR are identical for every paired task. Relative
to the wide binary gate, it improves SSIM by $+.026$ ($[+.018,+.035]$), while
its $+8.3$-point L-ESR difference is unresolved ($[0,+25.0]$). Because only
four of 12 concepts are detected and the cases target clear attachment
directions, this is evidence that asymmetric Permit can reduce collateral
freedom without lowering matched edit success, not a general semantic-quality
benchmark.

\subsection{Flow-magnitude threshold}
\label{supp:flow-threshold}
PromptPack-Flow begins with FlowEdit's source-to-target difference field
$V_s^\Delta$. The spatial gate $G_s$ removes updates outside the current edit
support, the magnitude threshold $\tau$ suppresses weak retained vectors, and
the boost $b$ scales the vectors that remain. This ablation asks how $\tau$
changes the balance between source preservation and edit realization:
\begin{equation}
\widetilde V_s^\Delta(u)=b\,G_s(u)\,\mathbb{1}\!\left[\lVert V_s^\Delta(u)\rVert_2\geq\tau\right]V_s^\Delta(u).
\end{equation}
The threshold, boost, and component sweeps below use saved pre-correction outputs with YOLO-World confidence $0.2$ and are retained as diagnostic trends. The K/P/M screen instead uses final corrected outputs and the main-paper confidence $0.1$.
Increasing $\tau$ suppresses more of the difference field, which raises source similarity but reduces the auxiliary detection rate (Tab.~\ref{tab:threshold_ablation}, Fig.~\ref{fig:threshold_tradeoff}). We select $\tau=4$ as an intermediate operating point rather than an optimum for either metric.

\begin{table}[h!]
\centering
\small
\setlength{\tabcolsep}{4pt}
\begin{tabular}{c|c|c}
\toprule
\textbf{Threshold} & \textbf{Detection (\%)} & \textbf{SSIM} \\
\midrule
0 & \textbf{43.1}$\pm$7.6 & 0.8058$\pm$0.0211\\
2 & 41.9$\pm$7.6 & 0.8118$\pm$0.0209\\
4 & 40.6$\pm$7.5 & 0.8508$\pm$0.0173\\
6 & 35.0$\pm$7.3 & 0.8987$\pm$0.0128\\
8 & 28.1$\pm$6.9 & 0.9313$\pm$0.0097\\
12 & 18.1$\pm$5.9 & 0.9541$\pm$0.0068\\
14 & 13.8$\pm$5.3 & \textbf{0.9573}$\pm$0.0068\\
\bottomrule
\end{tabular}
\caption{\textbf{Threshold ablation.} Detection is the fraction of 160 target concepts found by YOLO-World at confidence $0.2$ on pre-correction outputs. Intervals are 95\% confidence intervals; SSIM is measured across 80 images.}
\label{tab:threshold_ablation}
\end{table}

\begin{figure}[h!]
\centering
\includegraphics[width=\linewidth]{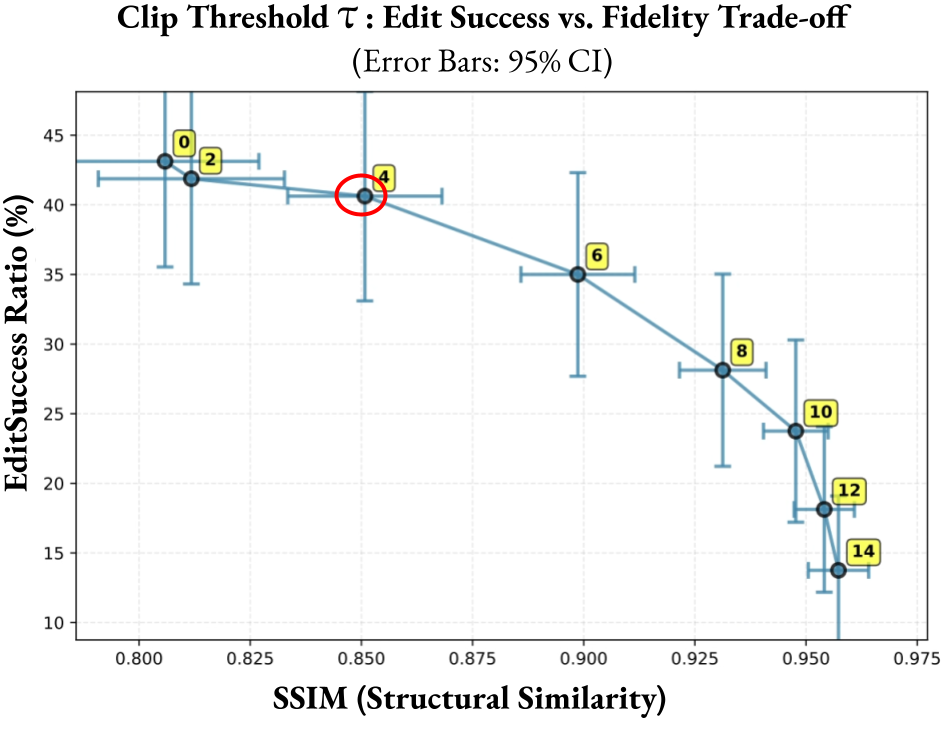}
\caption{Higher thresholds $\tau$ preserve more of the source but produce fewer detected target concepts in the auxiliary pre-correction sweep. Error bars are 95\% confidence intervals.}
\label{fig:threshold_tradeoff}
\end{figure}

\newpage
\vspace*{-0.05cm}
\subsection{Flow-boost factor}
\label{supp:flow-boost}
After spatial and magnitude filtering, the boost factor $b$ uniformly scales
the retained difference-field vectors. This sweep asks how much amplification
is needed to realize requested concepts before stronger updates primarily cause
source drift.

Increasing $b$ from 1.0 to 1.6 raises the auxiliary detection rate while decreasing SSIM. Beyond $b=1.6$, detection no longer improves, indicating that larger updates do not reliably produce the requested object. We therefore use $b=1.6$ for the final quantitative FlowEdit operating points. The component-ablation screen instead fixes $b=1.3$ to compare removal effects at a moderate edit-strength operating point; the K/P/M screen uses the final FlowEdit FLUX setting, $b=1.6$.

\begin{table}[h!]
\centering
\small
\setlength{\tabcolsep}{4pt}
\begin{tabular}{c|c|c}
\toprule
\textbf{Boost value $b$} & \textbf{Detection (\%)} & \textbf{SSIM} \\
\midrule
1.0 & 35.6$\pm$7.3 & \textbf{0.9029}$\pm$0.0121 \\
1.2 & 40.0$\pm$7.5 & 0.8681$\pm$0.0158 \\
1.4 & 44.4$\pm$7.6 & 0.8324$\pm$0.0193 \\
1.6 & \textbf{50.0}$\pm$7.7 & 0.7934$\pm$0.0226 \\
1.8 & 45.0$\pm$7.6 & 0.7593$\pm$0.0261 \\
2.0 & 43.8$\pm$7.6 & 0.7246$\pm$0.0293 \\
2.2 & 45.6$\pm$7.6 & 0.6932$\pm$0.0322 \\
\bottomrule
\end{tabular}
\caption{\textbf{Boost-factor ablation.} Detection uses a 95\% Wilson interval over 160 target concepts at YOLO-World confidence $0.2$; SSIM uses a 95\% $t$-interval over 80 images.}
\label{tab:boost_ablation}
\end{table}

\begin{figure}[h!]
\centering
\includegraphics[width=\linewidth]{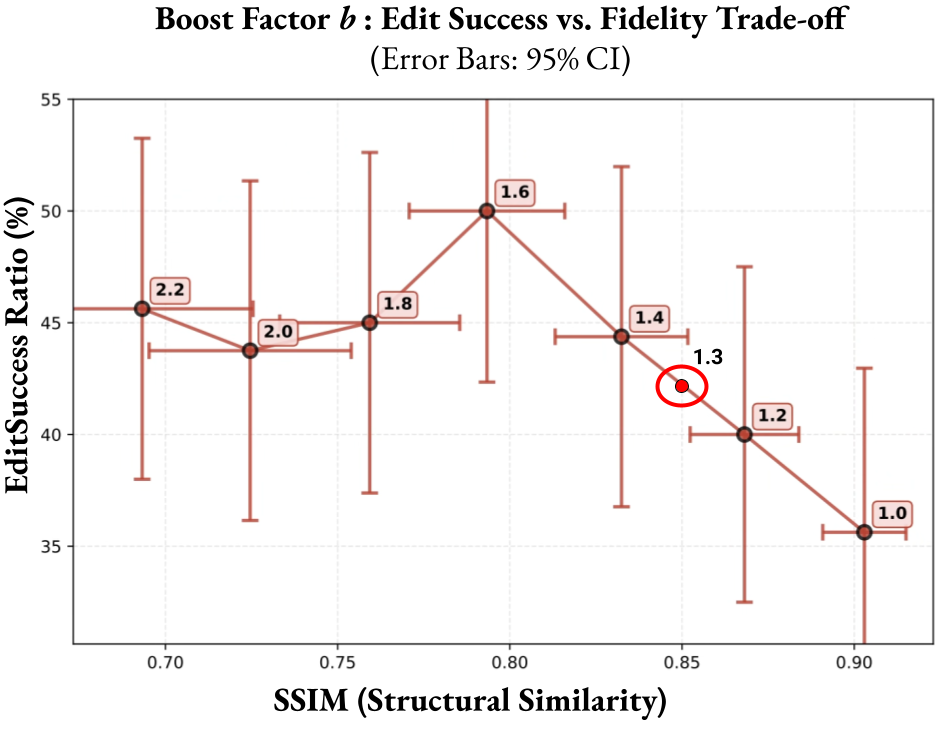}
\caption{Auxiliary pre-correction sweep of flow boost. Larger boosts increase target-concept detection up to $b=1.6$ while monotonically reducing source similarity.}
\label{fig:boost}
\end{figure}

\subsection{Component ablations}
\label{supp:component-ablations}
The preceding sweeps vary individual hyperparameters; this screen instead
removes one PromptPack mechanism at a time to test whether packing, routing,
spatial gating, thresholding, boosting, and compression correction make
distinct contributions. All variants are generated jointly from the same
inputs; the full reference configuration uses $(\tau,b)=(4,1.3)$, and each
ablation changes only the named component.
``No prompt packing'' retains the global target prompt but omits independent
regional embeddings; ``no attention redirection'' retains those embeddings but
removes their spatial masks.

\begin{table}[t]
\centering
\small
\setlength{\tabcolsep}{3.5pt}
\caption{PromptPack-Flow component and flow-localization ablations. Detection rates are percentages.}
\label{tab:ablation-results}
\begin{tabular}{lcccc}
\toprule
\textbf{Ablation} & \textbf{SSIM}$\uparrow$ & \textbf{LPIPS}$\downarrow$ & \textbf{ESR}$\uparrow$ & \textbf{CLIP}$\uparrow$ \\
\midrule
Full reference & .85 & .17 & 56.3 & .32 \\
No prompt packing & .86 & .16 & 49.4 & .32 \\
No attention redirection & .85 & .17 & 52.5 & .32 \\
No compression correction & .83 & .18 & 43.8 & .32 \\
No spatial gating & .83 & .21 & 50.0 & .32 \\
No magnitude threshold & .84 & .18 & 55.6 & .32 \\
No gate dilation & .86 & .17 & 53.1 & .32 \\
No flow boost & .89 & .13 & 48.1 & .32 \\
Threshold only & .83 & .20 & 49.4 & .32 \\
\bottomrule
\end{tabular}
\end{table}

Removing flow boost maximizes source similarity but lowers ESR; removing
prompt packing, redirection, dilation, or spatial gating also reduces edit
success. The reference is therefore a trade-off operating point rather than an
optimum for a single metric.

}

\clearpage
\onecolumn
\begingroup
\section{Additional Qualitative Results}
\label{supp:multi-turn}
Figures~\ref{fig:released-alternative-qualitative} and
\ref{fig:multiturn_full} provide the released-alternative and multi-turn
comparisons cited by the main paper. This section adds representative
PIE-Bench++ outputs, component-level failure modes, flow localization, and
limitations that are difficult to summarize with a single scalar metric.

\subsection{PIE-Bench++ visual comparisons}
\label{supp:piebench-visual-comparisons}
\vspace*{0.1cm}
    \centering
    \includegraphics[width=0.785\textwidth]{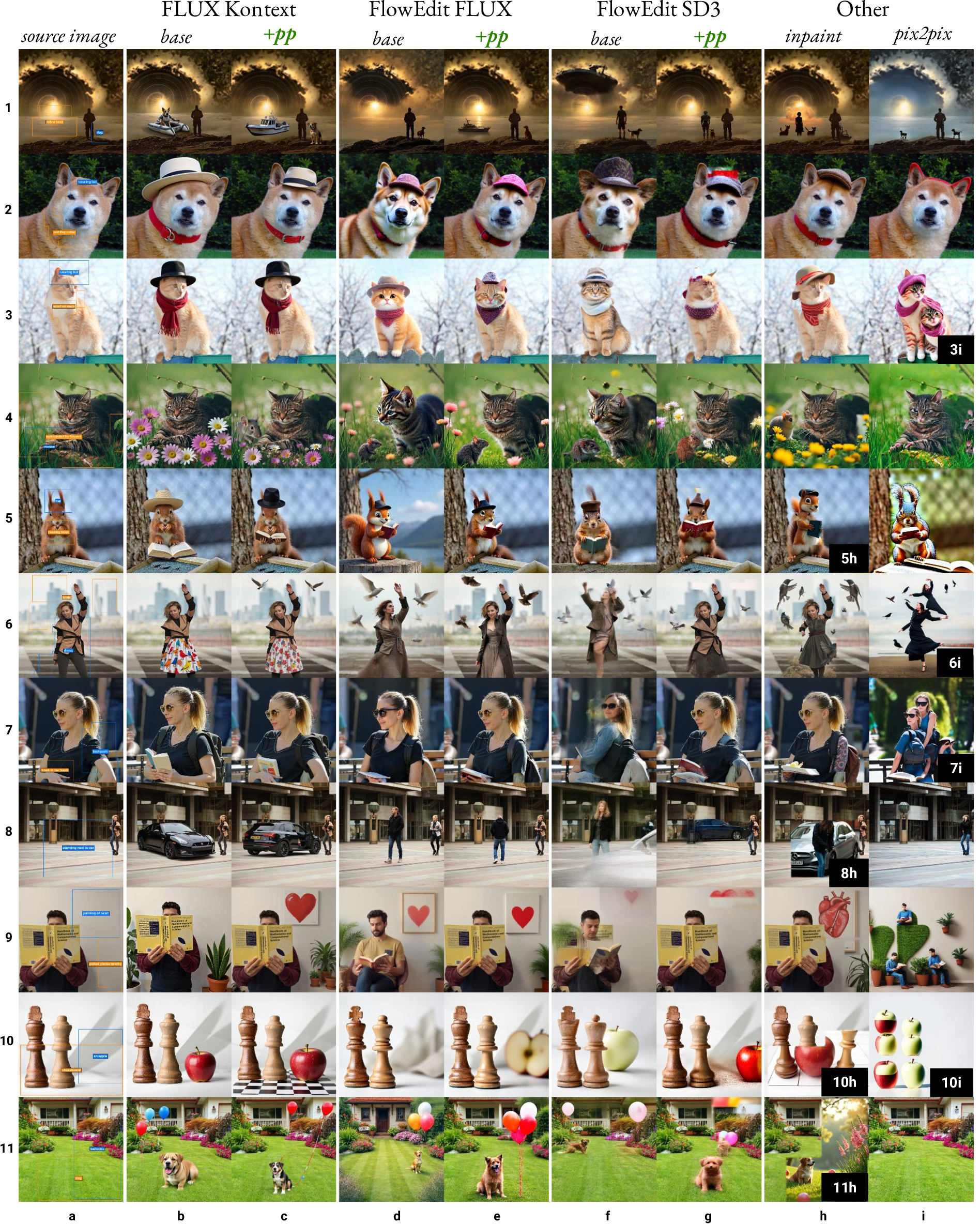}
  \captionof{figure}{\textbf{PIE-Bench++ examples.} Columns compare FLUX Kontext (b/c), FlowEdit FLUX (d/e), and FlowEdit SD3 (f/g) without/with PromptPack; FLUX inpainting (h) and InstructPix2Pix (i) are additional baselines. PromptPack often reduces background and identity drift, while the secondary baselines show boundary artifacts or duplicated objects in several examples. Prompts and masks are shown at left.}
    \label{fig:results_full}

\clearpage


\subsection{Component and flow-localization ablations}
\label{supp:qualitative-ablations}
\vspace*{-0.2cm}
\centering
\includegraphics[width=0.85\linewidth]{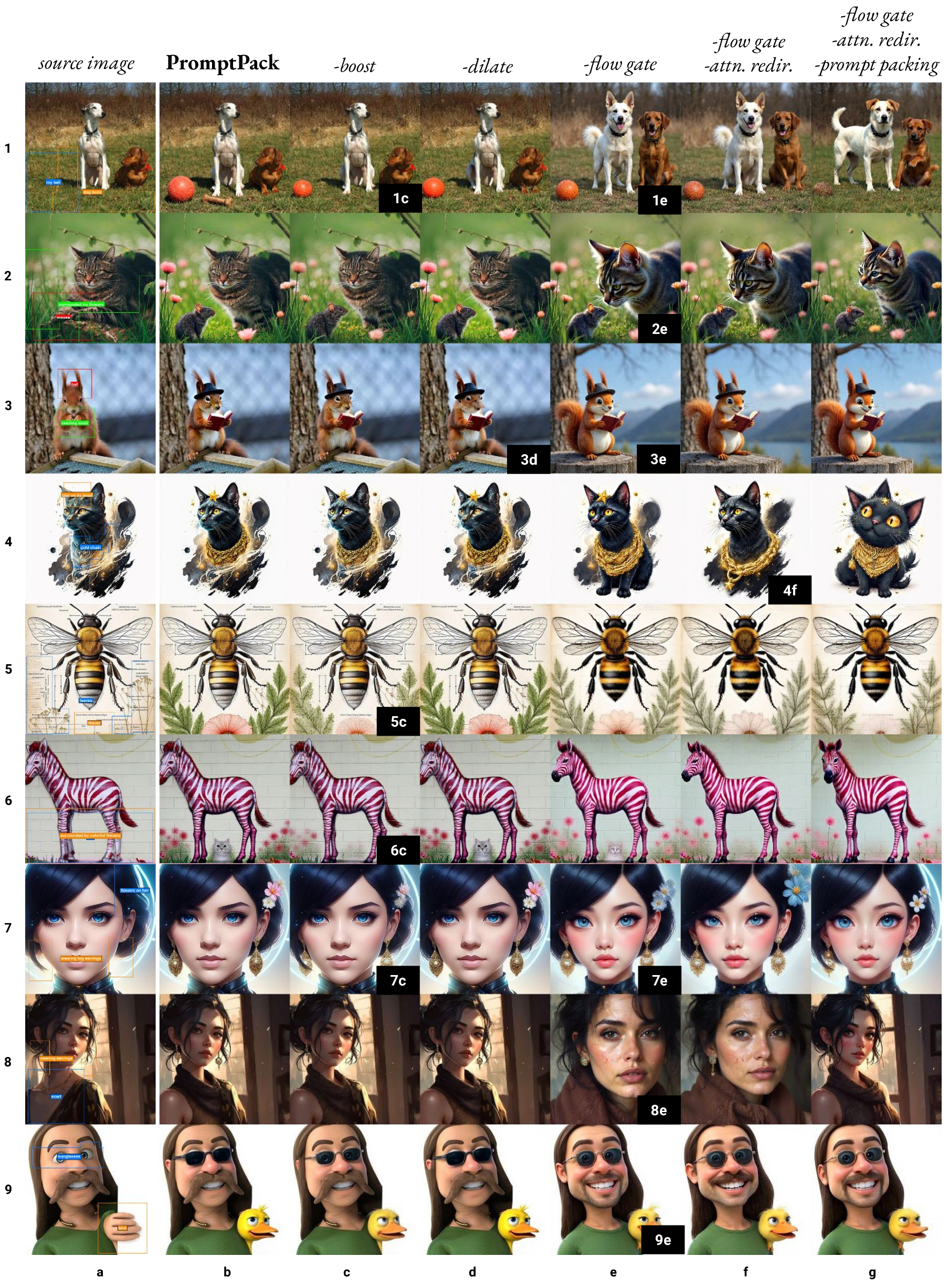}
\captionof{figure}{\textbf{General ablations.} Removing flow boosting (c) occasionally leads to edit concepts failing to generate (1c loses dog bone, 6c loses cat) or reduces the strength of generated edits (5c, 7c). Removing spatial edit gate dilation (d) causes visible edit integration artifacts (3d). Removing our entire flow gate mechanism (e) causes loss of subject identity (1e, 2e, 3e, 7e, 8e, 9e). Removing flow gating and attention redirection (f) also causes edit concepts to drop (4f).}
\label{fig:ablations}

\clearpage

\subsubsection*{Flow localization}
\centering
\includegraphics[width=0.9\linewidth]{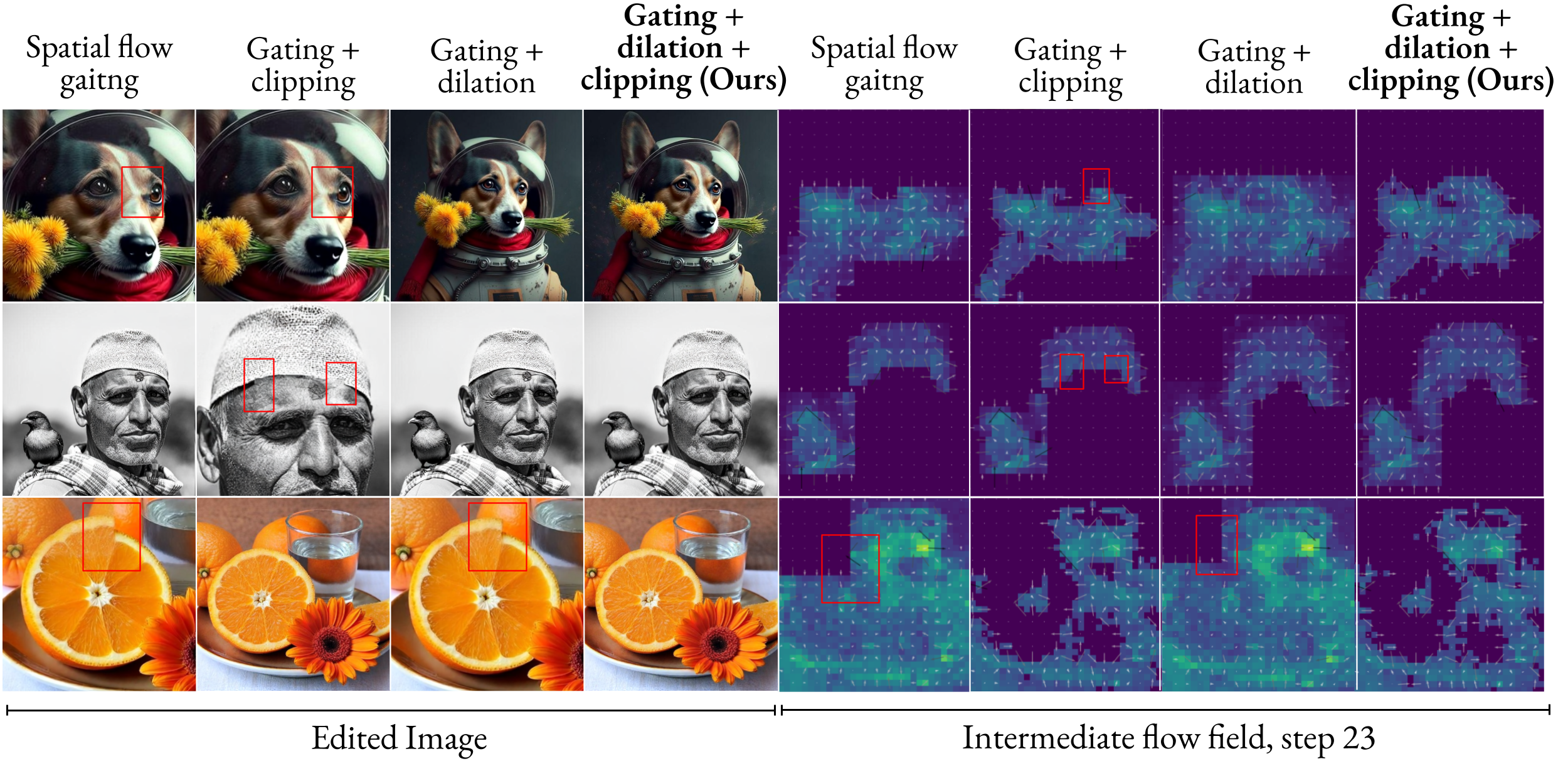}
\captionof{figure}{\textbf{Flow-localization ablations.} The left group shows enlarged output regions; the right group shows the corresponding flow-magnitude maps at iteration 23. Red boxes mark boundary artifacts and their associated flow discontinuities. Progressive dilation and magnitude thresholding concentrate changes within the allowed support and reduce these artifacts in the full configuration.}
\label{fig:localization}

\clearpage


\subsection{Failure cases}
\label{supp:failure-cases}
\centering
\vspace*{0.5cm}
\includegraphics[width=\linewidth]{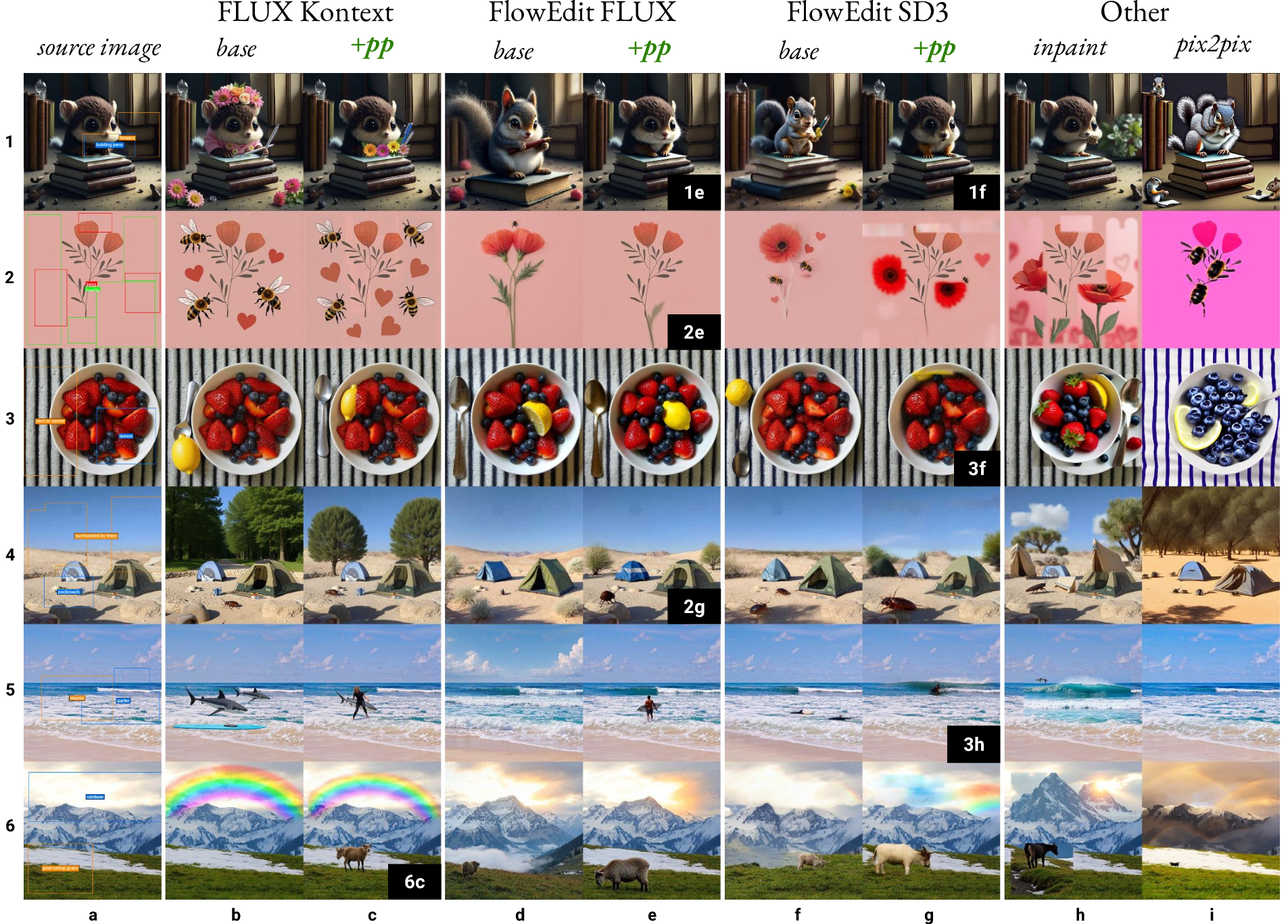}
\captionof{figure}{\textbf{Failure cases.} PromptPack can miss a target when its edit-role support or assignment mask excludes the location in which the base model attempts to form the concept. Expanding the \emph{Permit} region and assignment mask can mitigate this failure, at the cost of weaker localization. Conversely, dilation can leak content outside the original box; reducing the \emph{Permit} region or dilation extent can limit leakage. PromptPack can also preserve a base-model synthesis error, such as the two-headed goat in the final row.}
\label{fig:failures}

\endgroup
\clearpage
\twocolumn
\PromptPackDesignAblations

\clearpage
\section{Exploratory Video Extension}
\label{supp:video-extension}

\paragraph{Integration.} We apply the attention-redirection component from Sec.~\ref{subsection:attention_redirection} to VideoGrain~\cite{videograin}. SAM-Track~\cite{samtrack} supplies foreground masks across frames. These two examples are exploratory and are not part of the quantitative evaluation; they do not establish general video compatibility.

\begin{figure}[h!]
\centering
\includegraphics[width=\linewidth]{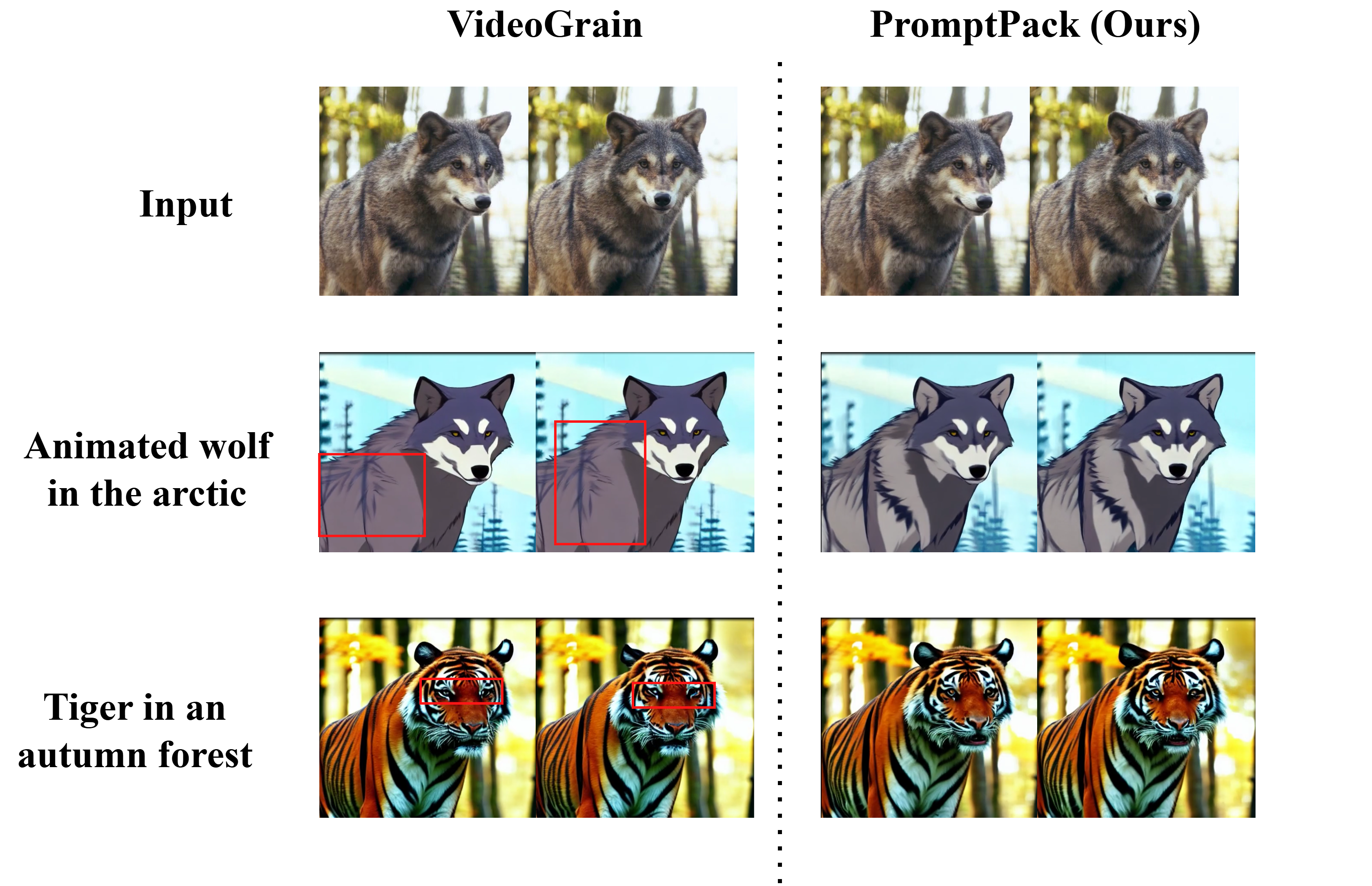}
\caption{Two qualitative VideoGrain comparisons. Red boxes mark foreground detail loss in the baseline: the wolf's multicolored fur becomes uniform, and the tiger's eyes blend into its face. Regional attention retains these details more clearly in the selected outputs.}
\label{fig:promptpack_vs_videograin}
\end{figure}

\paragraph{Observations.} Figure~\ref{fig:promptpack_vs_videograin} shows an animated-arctic edit and a tiger/autumn edit. In these selected examples, regional attention better separates foreground texture from background style. A systematic video study would still be needed to evaluate temporal consistency and generality.

\paragraph{Video availability.} This preprint includes only the
representative frames in Fig.~\ref{fig:promptpack_vs_videograin}; the
corresponding video clips are not included.

\end{document}